\documentclass{llncs}
\usepackage{graphicx}
\usepackage{comment}
\usepackage{amsmath,amssymb} 
\usepackage{color}

\usepackage[pagebackref=true,breaklinks=true,letterpaper=true,colorlinks,bookmarks=true]{hyperref}
\usepackage{siunitx}
\usepackage{wrapfig}
\usepackage{lscape}
\usepackage{rotating}
\usepackage{epstopdf}
\usepackage[nodisplayskipstretch]{setspace} 

\usepackage{booktabs}       
\usepackage{bm}       
\usepackage{gensymb}
\usepackage[shortlabels]{enumitem} 
\usepackage[noend,ruled,linesnumbered]{algorithm2e}
\graphicspath{{figure/}}
\usepackage{float}

\usepackage{pifont}

\newcommand{\cD}{\mathcal{D}}

\newcommand{\cO}{\mathcal{O}}
\newcommand{\cG}{\mathcal{G}}

\newcommand{\cV}{\mathcal{V}}
\newcommand{\cN}{\mathcal{N}}
\newcommand{\cE}{\mathcal{E}}

\newcommand{\cL}{\mathcal{L}}
 
\DeclareMathOperator*{\argmin}{arg\,min}

\newcommand{\etal}{\textit{et al}.}
\newcommand{\ie}{\textit{i}.\textit{e}.}
\newcommand{\eg}{\textit{e}.\textit{g}.}
\newcommand{\cf}{\textit{c}.\textit{f}.}
\newcommand\etc{etc{.\@}} 

\def\causal{\hbox{$\circ$}\kern-1.5pt\hbox{$\rightarrow$}}
\def\reversecausal{\hbox{$\leftarrow$}\kern-1.5pt\hbox{$\circ$}}

\usepackage{etoolbox}
\makeatletter
\patchcmd{\chapter}{\if@openright\cleardoublepage\else\clearpage\fi}{}{}{}
\makeatother

\emergencystretch=\maxdimen
\begin{document}

\pagestyle{headings}
\mainmatter
\def\ECCVSubNumber{4532}  

\title{NeuRoRA: Neural Robust Rotation Averaging}

\titlerunning{Neural robust rotation averaging}
%
\author{Pulak Purkait  \orcidID{0000-0003-0684-1209} \and
Tat-Jun Chin \orcidID{0000-0003-2423-9342} \and
Ian Reid \orcidID{0000-0001-7790-6423}}
\authorrunning{P. Purkait et al.}
%
\institute{ The University of Adelaide, Adelaide SA 5005, Australia 
\url{https://github.com/pulak09/NeuRoRA}}
\maketitle


\begin{abstract}
Multiple rotation averaging is an essential task for structure from motion, mapping, and robot navigation. 
The conventional methods for this task seek parameters of the absolute orientations that agree best with the observed noisy measurements according to a robust cost function. These robust cost functions are highly nonlinear and are designed based on certain assumptions about the noise and outlier distributions. In this work, we aim to build a neural network that learns the noise patterns from the data and predict/regress the model  parameters from the noisy relative orientations. The proposed network is a combination of two networks: (1) a view-graph cleaning network, which detects outlier edges in the view-graph and rectifies noisy measurements; and (2) a fine-tuning network, which fine-tunes an initialization of absolute orientations bootstrapped from the cleaned graph, in a single step. 
The proposed combined network is very fast, moreover, being trained on a large number of synthetic graphs, it is more accurate than the conventional iterative optimization methods. 
\end{abstract}


\section{Introduction}
Recently, we have witnessed a surge of interest in applying neural networks in various computer vision and robotics problems, such as, single-view depth estimation~\cite{garg2016unsupervised}, absolute pose regression~\cite{kendall2015posenet} and 3D point-cloud classification~\cite{qi2017pointnet}. However, we still rely on robust optimizations at different steps of geometric problems, for example, robot navigation and mapping. The reason is that neural networks have not yet proven to be effective in solving constrained optimization problems. Some classic examples of the test-time geometric optimization include rotation averaging~\cite{chatterjee2017robust,eriksson2018rotation,fredriksson2012simultaneous,hartley2013rotation,huynh2009metrics,wilson2016rotations} (a.k.a.  rotation  synchronization~\cite{arrigoni2018robust,singer2011angular,wang2013exact}), pose-graph optimization~\cite{kummerle2011g,tron2009distributed}, local bundle adjustment~\cite{mouragnon2009generic} and global structure from motion~\cite{wilson2014robust}. These optimization methods estimate the model parameters that agree best with the observed noisy measurements by minimizing a robust (typically non-convex) cost function. Often, these loss functions are designed based on certain assumptions about the sensor noise and outlier distributions. However, the observed noise distribution in a real-world application could be far from those assumptions.  A few such examples of noise patterns in real datasets are displayed in Figure~\ref{fig:statistics}. 
Furthermore the nature and the structure of the objective loss function is the same for different problem instances of a specific task. Nonetheless, existing methods optimize the loss function for each instance. Moreover, an optimization during test-time could be slow for a target task involving a large number of parameters, and often forestalls a real-time solution to the problem.  
\begin{figure}[H]
\includegraphics[width=1\textwidth]{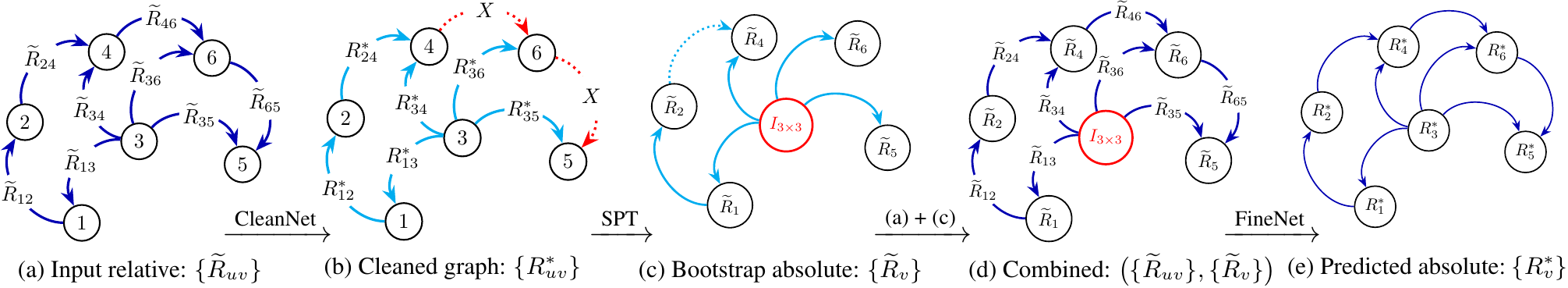} 
\caption{The proposed method NeuRoRA is a two-step approach: in the first step a graph-based network (CleanNet) is utilized to clean the view-graph by removing outliers and rectifying noisy measurements. An initialization from the cleaned view-graph, instantiated from a shortest path tree (SPT), is then further fine-tuned using a separate graph-based network (FineNet). The notations are outlined in Table~\ref{tab:notations}. 
}
\label{fig:token} 
\end{figure} 
In this work, with the advancement of machine learning, we address the following question: \emph{``can we learn the noise patterns in data, given thousands of different problem instances of a specific task, and regress the target parameters instead of optimizing them during test-time?''} The answer is affirmative for some specific applications, and we propose a learning framework that exceeds baseline optimization methods for a geometric problem. We choose \emph{multiple rotation averaging} (MRA) as a target application to validate our claim. 
   
In MRA, the task is to estimate the absolute orientations of cameras given some of their pairwise noisy relative orientations defined on a view-graph. There are a different number of cameras for each problem instance of MRA, and usually sparsely connected to each other. Further, the observed relative orientations are often corrupted by outliers. The conventional methods for this task~\cite{arrigoni2018robust,chatterjee2017robust,eriksson2018rotation,hartley2013rotation,wang2013exact} optimize the parameters of the absolute orientations of the cameras that are most compatible (up to a robust cost) with the observed noisy relative orientations.  

We propose a neural network for robust MRA. Our network is a combination of two simple four-layered message-passing neural networks defined on the view-graphs, 
summarized in Figure~\ref{fig:token}. 
We name our method \emph{Neural Robust Rotation Averaging}, which is abbreviated as {NeuRoRA} in the rest of the manuscript. 

\noindent {\bf Contribution and findings} 
\begin{itemize}[noitemsep,nolistsep,leftmargin=3.5mm,topsep=0ex,partopsep=0ex] 
    \item A graph-based neural network NeuRoRA is proposed as an alternative to conventional optimizations for MRA. 
    \item NeuRoRA requires \emph{no} explicit optimization at test-time and hence it is {\bf much faster} $(10-50\times$ on CPUs and $500-2000\times$ on GPUs) than the baselines. 
    \item The proposed NeuRoRA is {\bf more accurate} than the conventional optimization methods of MRA. The mean/median orientation error of the predicted absolute orientations by the proposed method is $1.45\degree / 0.74\degree$, compared to  $2.17\degree/1.25\degree$ by an optimization method~\cite{chatterjee2017robust}. 
    \item Being {\bf a small size} network, NeuRoRA is fast and can easily be deployed to real-world applications (network size $<0.5$Mb). 
\end{itemize}
\begin{figure}[H]
\includegraphics[width=1\textwidth]{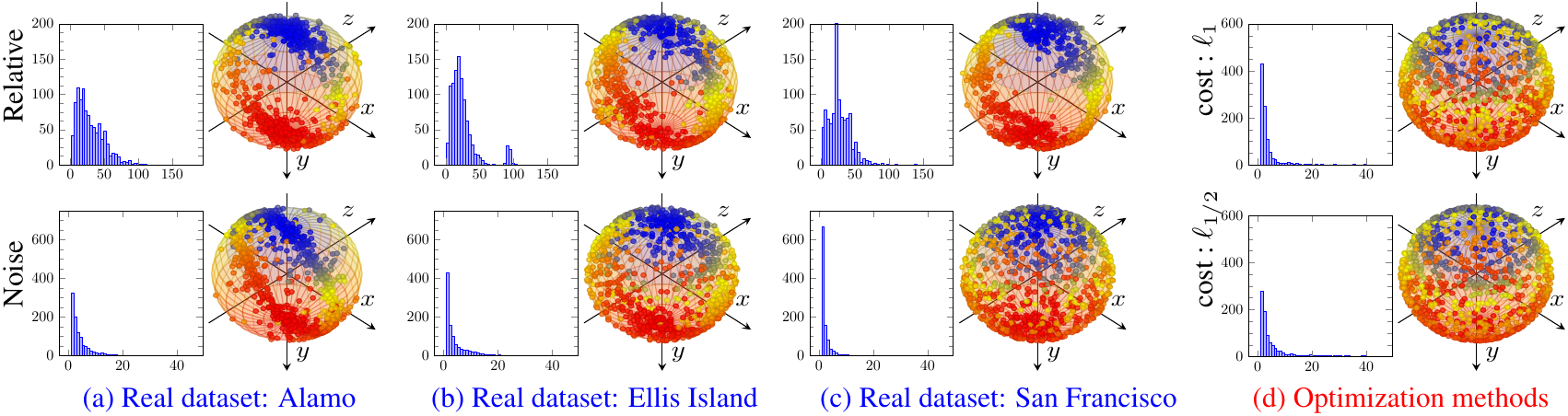}
 \caption{Here we qualitatively illustrate that the noise distribution in real data diverges considerably from the noise assumptions baked into most optimization methods. We plot the angle and axes of observed {\bf relative} orientations (first row) and the same of {\bf noise} (second row) in real datasets (for clarity only ${10^3}$ random samples) are displayed. The noise orientation is calculated from the ground-truth absolute and the observed relative orientations. The view-graphs of \textcolor{blue}{(a)-(b)} are shared by~\cite{wilson2014robust} and \textcolor{blue}{(c)} is shared by~\cite{crandall2011discrete}. We plotted histograms of the magnitudes of the angles (in degrees) and the axes of the orientations. Notice that the axes of the sampled relative and noise orientations 
 for the real data in (a)-(c) are not uniformly distributed on the unit ball. 
The sampled noise orientations (somewhat vertical axes) are far from the typical distribution assumptions regarded by optimization algorithms. Samples from such noise distributions ($\ell_1$~\cite{wang2013exact} and $\ell_{1/2}$~\cite{chatterjee2017robust}) are shown in \textcolor{red}{(d)}. }
 \label{fig:statistics}
\end{figure}
\section{Related works}\label{sec:relative}
We separate the related methods into two separate sections---{\bf (i)} learning based methods as an alternative to optimizations, {\bf (ii)} relevant optimizations specific to MRA, and {\bf (iii)} other related optimization methods. 

\noindent {\bf (i) Learning to optimize}
A neural network is proposed as an alternative to non-linear least square optimizations in~\cite{clark2018learning} for camera tracking and mapping. It exploits the least square structure of the problem and uses a recurrent network to compute updated steps of the optimization variables. 
In a similar direction,~\cite{lv2019taking} relaxes the assumptions made by inverse compositional algorithms for dense  image alignment by incorporating data-driven priors.  \cite{tang2018ba} proposes a bundle adjustment layer that learns to  predict the dampening parameter of the Levenberg-Marquardt algorithm to optimize depth and camera parameters. In contrast to the  direct optimization-based methods that explicitly use regularizers to solve an ill-posed problem, \cite{adler2017solving} implicitly learn the prior through a data-driven method. Aoki~\etal~\cite{aoki2019pointnetlk} proposed an iterative algorithm based on PointNet~\cite{qi2017pointnet} for point-cloud registration as an alternative to direct optimization. Learning to predict an approximate solution to combinatorial optimization problem over graphs, \eg~minimum vertex cover, traveling salesman problem, \etc,~is proposed in~\cite{khalil2017learning}. Learning methods to optimize general black-box functions~\cite{chen2017learning} have also received a lot of attention recently. These conventional learning-based methods are tailored to some specific problems, where in this work, we are interested in an alternative learning-based solution for geometric problems, \eg~SFM.

\noindent  {\bf (ii) Robust optimization for rotation averaging}
MRA was first introduced in~\cite{govindu2001combining} where a linear solution was proposed using quaternion averaging and later in~\cite{govindu2004lie} using Lie group based averaging. The solutions were non-robust in both the cases. 
Recently, there has been progress in designing robust algorithms~\cite{chatterjee2013efficient,hartley2011l1} for rotation averaging. 
Most of the algorithms are based on iterative methods for optimizing a robust loss function. 
MRA is also exploited using sparse matrix decomposition, for example~\cite{arrigoni2018robust,wang2013exact}. 
The state of the art methods are listed below: 
\begin{itemize}[noitemsep,nolistsep,leftmargin=4.5mm,topsep=0ex,partopsep=0ex]
\item Chatterjee and Govindu~\cite{chatterjee2017robust} fine-tune an initialization by first performing an iterative $\ell_1$ minimization, followed by another iterative reweighted least squares with a more robust loss function $\ell_\frac{1}{2}$. 
\item Hartley~\etal~\cite{hartley2011l1} propose a straight-forward method. It fine-tunes an initialization by the Weiszfeld algorithm of $\ell_1$ averaging~\cite{chandrasekaran1989open}. At every iteration, the absolute orientations of each camera are updated by the median of those computed from its neighbors. 
\item  Arrigoni~\etal~\cite{arrigoni2018robust} formulate the problem as a low-rank and sparse matrix decomposition and utilizes existing decomposition algorithms that caters for missing data, outliers and noise in the pairwise observations. 
\item  Wang~\etal~\cite{wang2013exact} employ the alternating direction method to minimize a robust cost function involving the sum of unsquared deviations. 
\end{itemize} 
Rotation averaging is surveyed recently in a vast amount of literature~\cite{arrigoni2020synchronization,carlone2015initialization,ozyecsil2017survey,tron2016survey}.

\noindent  {\bf (iii) Other related optimization methods}
DISCO~\cite{crandall2011discrete} employs a two-step approach. In the first step, 
a loopy belief propagation is used for an initial estimation of the absolute orientations which are fine-tuned by Levenberg-Marquardt method in the second step. The problem of detecting outliers in the view-graph has been extensively studied in the literature~\cite{enqvist2011non,govindu2006robustness,jiang2013global,moulon2013global,shah2018view,zach2010disambiguating}. Optimizing/cleaning the view-graph for sfm~also proposed in \cite{guibas2019condition,shen2016graph,sweeney2015optimizing}. 

Huang~\etal~\cite{huang2019learning} proposed a neural network solving the pairwise matching problem (\cf~page~2, 2nd col) accurately. The key component of~\cite{huang2019learning} is a network that takes two 3D scans and a relative transformation between them as input and outputs a score indicating the goodness of the scan alignment which is iteratively employed to fine-tune the absolute pose. Therefore, \cite{huang2019learning} is only valid (and tailored) for alignments of multiple scans. 


\section{Multiple rotation averaging}
Consider $N$ cameras with $M$ pairwise relative orientation measurements forming a directed view-graph $\cG = (\cV, \cE)$. A vertex $\cV_{v} \in \cV$ corresponds to the absolute orientation $\widehat R_{v}$ (to be estimated) of the $v$th camera and an edge $\cE_{uv} \in \cE$ corresponds to the observed relative orientation $\widetilde{R}_{uv}$ from $u$th camera to $v$th  camera.  
Conventionally, the task is to estimate the absolute orientations $\{\widehat R_{v}\}$, with  respect to a global reference of orientations, such that the  estimated orientations are most consistent with the observed noisy relative orientation measurements, \ie~$\widetilde{R}_{uv} \approx \widehat R_{v} \widehat R_{u}^{-1}, \forall \cE_{uv} \in \cE $. Further, the observed measurements are corrupted by  outliers, \ie~some of the orientations $\widetilde{R}_{uv}$ are far from $ \widehat R_{v} \widehat R_{u}^{-1}$. Conventionally, the solution is obtained by minimizing a robust cost function that penalizes the  discrepancy between observed noisy relative orientations $\{\widetilde{R}_{uv}\}$ and the estimated relative orientations $\{R_{uv}^\ast\} := \{R_{v}^\ast {R_{u}^\ast}^{-1}\}$. The corresponding optimization problem can then be expressed as 
\begin{equation}
\argmin_{\{R_{v}^\ast\}} \sum_{\cE_{uv} \in \cE}\rho\Big(d\big(R_{uv}^\ast, \widetilde R_{uv}\big)\Big)
\label{eq:compatibility}
\end{equation}
where $\rho(.)$ is a robust cost and $d(.,.)$ is a distance measure between the orientations. 
The nature of the above optimization is a typical complex multi-variable nonlinear optimization problem with thousands of variables (for thousands of cameras) and there seems to be no direct method (closed-form solution) minimizing the above cost even without outliers~\cite{hartley2013rotation}. 

\noindent {\bf The choice of distance measure $d(\widetilde R, R)$} 
There are three commonly used distance measurements in the rotation group SO(3): (i) the geodesic or angle metric $d_\theta = \angle(\widetilde R, R)$, (ii) the chordal metric $d_{C} = \| \widetilde R - R \|_F$ and (iii) the quaternion metric $d_{Q} = \min{ \{\| q_{\widetilde R} - q_{R} \|, \| q_{\widetilde R} + q_{R} \|\}}$ where $q_{R}$ and $q_{\widetilde R}$ are quaternion representations of $R$ and $\widetilde R$ respectively, and $\|.\|_F$ is the Frobenius norm. 
The metrics $d_C$ and $d_{Q}$ are proven to be $2\sqrt{2}\sin(d_\theta/2)$ and $2\sin(d_\theta/4)$ respectively~\cite{hartley2013rotation}, thus, all the metrics are the same to the first order. In our implementation, we employ the quaternion representations (with non-negative scalars). 

\noindent {\bf The choice of robust cost $\rho(.)$} 
In practical applications, \eg~robot navigation, the agent usually ends up with some corrupt measurements (outliers), due to symmetric and repetitive man-made structures, in addition to the sensor noise. To estimate the absolute orientations of the cameras that are immune to those outliers, the conventional methods optimize a robust cost $\rho(.)$ as discussed above. An exhaustive list of such robust functions can be found  in~\cite{chatterjee2017robust}. 

The noise and outliers in the observed relative orientations is assumed to follow some distributions subject to the cost function  with mean identity orientation~\cite{hartley2013rotation,wang2013exact}. However, in real data, we observe very different noise distributions and a few such examples are shown in Figure~\ref{fig:statistics}. Further, optical axis of most of the cameras are horizontal and hence the axes of the relative orientations are vertical. By training a neural network to perform the task, our aim is for the neural network to capture these patterns while predicting the absolute orientations.  


\section{Learning to predict absolute orientations}
Let $\cD := \{\cG\}$ be a dataset of ground-truth view-graphs. Each view-graph $\cG:=(\cV, \cE)$ contains a noisy relative orientation measurement $\widetilde R_{uv}$ for each edge $\cE_{uv} \in \cE$ and a ground-truth absolute orientation $ \widehat R_{v}$ for each camera $\cV_v \in \cV$. The desired neural network learns a mapping $\Phi$ that takes noisy relative measurements $\{\widetilde R_{uv}\}$ as input and predicts the absolute orientations $\{R^{\Phi}_{v}\}:=\Phi\Big(\{\widetilde R_{uv}\};\;\Theta\Big)$ as output, where $\Theta$ is the set of network parameters. To train the parameters of such network, one could minimize the discrepancy between the ground-truth $\widehat R_{uv}:= \widehat R_{v} \widehat R_{u}^{-1}$ and the estimated $R_{uv}^{\Phi} := R_{v}^{\Phi} {R_{u}^{\Phi}}^{-1}$ relative orientations  (\emph{cf.} equation~\eqref{eq:compatibility}), \ie 
 \begin{equation}
\argmin_{\Theta} \sum_{\cG \in \cD}\sum_{\cE_{uv} \in \cE} d\big(R_{uv}^{\Phi}, \widehat R_{uv}\big)
\label{eq:newloss}
\end{equation}
In contrast to~\eqref{eq:compatibility}, where conventional methods optimize the orientation parameters for each instance of the view-graph $\cG \in \cD$, here in~\eqref{eq:newloss}, the network parameters are optimized during training that learn the mapping $\Phi$ effectively from observed relative orientations $\{\widetilde R_{uv}\}$ to the target absolute orientations $\{\widehat R_{v}\}$, \ie~$\{\widehat R_{v}\} \approx \Phi\Big(\{\widetilde R_{uv}\};\;\Theta\Big)$ over the entire dataset of view-graphs $\cD$. 

\noindent {\bf Direct training of $\Phi$ and gauge freedom}
For an arbitrary orientation $R$, 
\begin{equation}
R_{uv}^{\ast} := R_{v}^{\ast} {R_{u}^{\ast}}^{-1}  = ( R_{v}^{\ast}R){(R_{u}^{\ast}R)}^{-1},\;\;\; \forall \cE_{uv} \in \cE
\end{equation}
Therefore, $\{R^{\ast}_{v}\}$ and $\{R^{\ast}_{v} R\}$ essentially represent the same solution to the MRA problem~\eqref{eq:compatibility} and there is a gauge freedom of degree $3$. The mapping $\Phi$ is thus one-to-many as $\{R^{\Phi}_{v}\}$ and $\{R^{\Phi}_{v} R\}$ correspond to the same cost~\eqref{eq:newloss}. This gauge freedom makes it difficult to train such a network. Further, one could choose a direct cost (no associated gauge freedom) to learn an one-to-one mapping $\Phi$, 
\eg 
 \begin{equation}
\argmin_{\Theta} \sum_{\cG \in \cD} \sum_{\cV_v \in \cV} d\big( R_{v}^{\Phi}, \widehat R_{v} \big) 
\label{eq:newloss22}
\end{equation}
where the reference orientation is fixed according to the ground-truth. Again, $\{\widehat R_{v}\}$ and $\{\widehat R_{v} R\}$ represent the same ground-truth where the reference orientations are fixed at different directions. One could fix the issue by fixing the reference orientation to the orientation of the first camera in all the view-graphs in $\cD$. 
However, in a graph (set representation), the nodes are permutation invariant. Thus the choice of the first camera, and hence the reference orientation, is arbitrary. 
Therefore, one needs to pass the reference orientation or the index of the first camera (possibly via a binary encoding) to the network as an additional input to be able to train such a network. However, we employ an alternative  strategy adopted from the conventional optimization methods~\cite{chatterjee2017robust,hartley2013rotation}, \ie~initialize a solution of the absolute orientations under a fixed reference and pass the initialization to the network to fine-tune the solution.   
The network gets the reference orientation as an additional input via initialization (see Figure~\ref{fig:token}(d)) and regress the parameters, \ie~$\{\widehat R_{v} \}\approx \Phi\Big(\{\widetilde R_{uv}\}, \{\widetilde R_v\};\;\Theta\Big)$. Further, we train the network by minimizing a combined cost where the 1st term~\eqref{eq:newloss} enforces the consistency over the entire graph and the 2nd term~\eqref{eq:newloss22} enforces a unique solution, \ie  
 \begin{equation}
\argmin_{\Theta} \sum_{\cG \in \cD} \Big( \sum_{\cE_{uv} \in \cE} d\big(R_{uv}^{\Phi}, \widehat R_{uv}\big) + \beta \sum_{\cV_v \in \cV} d\big( R_{v}^{\Phi}, \widehat R_{v} \big)\Big) 
\label{eq:newloss2}
\end{equation}
where $\beta$ is a weight parameter. Note that the reference orientation are now fixed at the orientation of a certain camera $c$ in the initialization  $\{\widetilde R_{v}\}$ as well as in the ground-truth absolute orientations $\{ \widehat R_{v} \}$. Although, the choice of $c$ is not critical, in practice, the camera $c$ with most neighboring cameras is chosen as the reference, \ie~$\widetilde R_{c}$ = $\widehat R_{c}$ = $I_{3\times 3}$.  

The above mapping $\Phi$ is now one-to-one. However, it requires an initialization $\{\widetilde R_{v}\}$ as an additional input. Conventional methods initialize the absolute orientations using a spanning tree of the view graph. However even a single outlier in that spanning tree can lead to a very poor initialization, so it is very important to identify these outliers beforehand. Further, noise in the relative orientation along each edge of the spanning tree will also propagate at the subsequent nodes while computing the initial absolute orientations. Thus, we first clean the view-graph by removing the outliers and rectifying the noisy measurements, and then bootstrap an initialization from the cleaned view-graph. 

\noindent {\bf Cleaning the view-graph}
Given the local structure in the view-graph, \ie~measurements of all the edges that the pair of adjacent nodes $\{\cV_{u}, \cV_{v}\}$ are connected to (and possibly subsequent edges), an outlier edge $\cE_{uv}$ can be detected. To be specific, chaining the relative orientations along a cycle in the local structure of the view-graph forms an orientation close to the identity orientation and an indication of an outlier in the cycle otherwise. The presence of an outliers in multiple such cycles through the current edge indicates that the edge to be an outlier. 
Instead of designing such explicit algorithms, we use another neural network to clean the graph. The proposed method can be summarized as follows:
\begin{itemize}[noitemsep,nolistsep,leftmargin=4.5mm,topsep=0ex,partopsep=0ex] 
\item A graph-based network is employed to clean the view-graph by removing outlier measurements and rectifying noisy measurements (see  Section~\ref{sec:cleannet}). 
\item The cleaned view-graph is then utilized to initialize the absolute orientations (see Section~\ref{sec:init}). 
\item The initialization is fine-tuned using a separate network (see Section~\ref{sec:finetune}). 
\end{itemize}
For clarity of the rest of the paper, the notations are outlined in Table~\ref{tab:notations}. 

\subsection{The network design choice}\label{proposednet}
\begin{figure}[H] 
\centering
\includegraphics[width=0.9\textwidth]{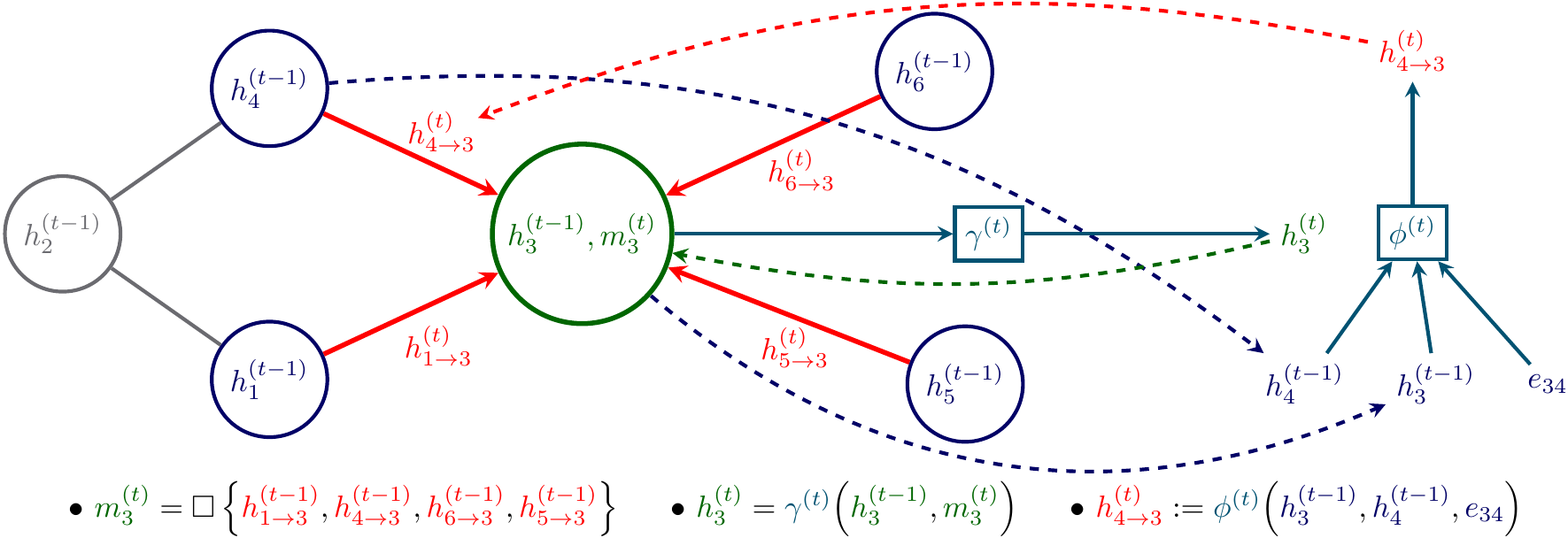}
\caption{An illustration of computing next level features of a message-passing network.} 
\label{fig:mp}
\end{figure}

Generalizing convolution operators to irregular domains, such as graphs, is typically expressed as neighborhood aggregation or a message-passing scheme. The proposed network is built using such Message-Passing Neural Networks (MPNN)~\cite{gilmer2017neural},  directly operating  on view-graphs $\cG$. 
A MPNN is defined in terms of message functions $m_v^{(t)}$ and update functions $\gamma^{(t)}$ that run for $T$ time-steps (layers). At each time-step, the hidden state $h_v^{(t)}$ at each node (feature) in the graph is updated according to   
\begin{equation}
h_v^{(t)} = \gamma^{(t)}\big( h_v^{(t - 1 )}, m_v^{(t)} \big)
\label{eq:gamma}
\end{equation}
where $m_v^{(t)}$ is the condensed message at node $v$, coming from the neighboring nodes $u \in \cN_v$, and can be expressed as follows:  
\begin{equation}
m_v^{(t)} = \square_{\cV_{u} \in \cN_v} \phi^{(t)}\big( h_v^{(t - 1 )}, h_u^{(t - 1 )}, e_{uv}\big) 
\label{eq:phi}
\end{equation}
where $\square$ denotes a differentiable, permutation invariant symmetric function, \eg~\emph{mean}, \emph{soft-max}, \etc; $\gamma^{(t)}$ and $\phi^{(t)}$ are concatenation operations followed by $1$-D convolutions and ReLUs; $e_{uv}$ is the edge feature of the edge $\cE_{uv}$, $h^{(t)}_{u\rightarrow v}:=\phi^{(t)}\big( h_v^{(t - 1 )}, h_u^{(t - 1 )}, e_{uv}\big)$ is the accumulated message for the edge $\cE_{uv}$ at time-step $(t)$;  and $\cN_v$ is the set of all neighboring cameras connected to $\cV_v$. A diagram of the elements involved in computing the next-level features is shown in Figure~\ref{fig:mp}. 

\begin{table}[H] \setlength{\tabcolsep}{0.5pt}
\caption{The notations and symbols used in the manuscript}
\scriptsize 
\begin{tabular}{@{\hspace{0.3em}}lcp{4.6cm}@{\hspace{0.55em}}|@{\hspace{0.55em}}lcp{4.2cm}}
\toprule
\multicolumn{6}{l}{\bf{Orientation parameters  in the view-graph}} \\ 
\midrule 
$\widetilde R_{uv}$$|$$R_{uv}^{\ast}$ & $:$ & Observed$|$Noise-rectified relative &  $\widetilde R_{v}$$|$$R_v^{\Phi}|R_{v}^{\ast}$ & $:$ & Initial$|$Refined$|$predicted absolute   \\
$\widehat R_{uv}$$|$$\widehat R_{v}$ & $:$ & Ground-truth relative$|$absolute  & $\{R_{uv}\}$$|$$\{ R_{v} \}$ & $:$ & Set of all relative$|$absolute  \\ 
\midrule
\multicolumn{6}{l}{\bf{The network parameters and symbols}} \\ 
\midrule 
$\alpha_{uv}^{\ast} $$|$$\widehat \alpha_{uv} $ & $:$ & Predicted$|$Ground-truth outlier-score &
$h_v^{(t)}$$|$$m_v^{(t)} $  & $:$ & Features$|$Message at node $v$  \\ 
$\phi^{(t)}$$|$$\gamma^{(t)}$ & $:$ & Feature update$|$Accumulated message &   
$lp_1, lp_2, lp_3$ & $:$ & Single layers of linear perceptrons \\ \bottomrule 
\end{tabular}
\label{tab:notations}
\end{table}
\subsection{View-graph cleaning network}\label{sec:cleannet}
The view-graph cleaning network (CleanNet) is built on a MPNN.  The input to CleanNet is a noisy view-graph and the output is a clean one, \ie~the network takes noisy relative orientations $\widetilde R_{uv}$ as the edge features $e_{uv}$ and predicts the noise-rectified relative orientations $R_{uv}^{\ast}$ from the accumulated message $h^{(T)}_{u\rightarrow v}:=\phi^{(T)}\big( h_v^{(T - 1 )}, h_u^{(T - 1 )}, e_{uv}\big)$ at the last layer. It also predicts a score $\alpha_{uv}^{\ast}$ depicting the probability of the edge $\cE_{uv}$ to be an outlier. \ie~
\begin{align}
 R_{uv}^{\ast} =  lp_1\Big(h^{(T)}_{u\rightarrow v}\Big) \star \widetilde R_{uv}   ~~\text{and}~~
 \alpha_{uv}^{\ast} = lp_2\Big(h^{(T)}_{u\rightarrow v}\Big) \label{eq:lp12}
\end{align}
where $lp_1(.)$ and $lp_2(.)$ are single-layered linear perceptrons that map the accumulated messages to the edge noise orientation and outlier score respectively.  $'\star'$ is the matrix multiplication. The hidden states are initialized by null vectors, \ie~$h_v^{(0)} = \emptyset$. Note that instead of directly estimating the rectified orientations, we predict the noise in the relative orientation measurements, which are then multiplied to obtain the rectified orientations. The loss is chosen as the weighted combination of mean orientation error  $\cL_{mre}$ of the rectified $ R_{uv}^{\ast}$  and ground-truth $\widehat R_{uv} := \widehat R_{v} \widehat R_{u}^{-1}$ relative orientations, and mean binary cross-entropy error $\cL_{bce}$ of the predicted $ \alpha_{uv}^{\ast}$ and the ground-truth outlier score $\widehat \alpha_{uv}$, \ie 
\begin{equation}
\cL = \sum_{\cG \in \cD} \sum_{\cE_{uv} \in \cE} \Big(  \cL_{mre}\big(R_{uv}^{\ast}, \widehat R_{uv}\big) + \lambda \cL_{bce}\big(  \alpha_{uv}^{\ast}, \widehat\alpha_{uv}\big) \Big) 
\end{equation}
where $\lambda$ is a weight parameter (fixed as $\lambda=10$). 
We formulate the orientations using unit quaternions 
and the predictions are normalized accordingly. The error in the prediction is also normalized by the degree of the node, \ie 
\begin{equation}
\cL_{mre}\big(R_{uv}^{\ast}, \widehat R_{uv} \big) = \frac{1}{|\cN_v||\cN_u|} d_{Q}\big(\frac{R_{uv}^{\ast}}{\|R_{uv}^{\ast}\|_2}, \widehat R_{uv}\big)
\label{eq:quat}
\end{equation}
Experimentally, we observed the above loss produces superior performance than the standard discrepancy loss~\eqref{eq:newloss}. Note that the ground-truth outlier score $\widehat \alpha_{uv}$ is generated based on the amount of noise in the relative orientations. Specifically, if the amount of noise in the relative orientation $\widetilde R_{v}^{-1} \widehat R_{uv} \widetilde R_{u} > 20\degree$,  the ground-truth edge label is assigned as an outlier, \ie~$\widehat \alpha_{uv} = 1$ and $\widehat\alpha_{uv} = 0$ otherwise. 
An edge $\cE_{uv}$ is marked as an outlier edge if the predicted outlier score $ \alpha_{uv}^{\ast}$ is greater than a predefined threshold $\epsilon$. In all of our experiments, we choose the threshold $\epsilon = 0.75$\footnote{The choice is not critical in the range $\epsilon \in [0.35, 0.8]$.}. A cleaned view-graph $ \cG^{\ast}$ is then generated by removing outlier edges from $\cG$ and replacing  noisy relative orientations $\widetilde R_{uv}$ by the rectified orientations $ R_{uv}^{\ast}$.  Note that the cleaned graph $\cG^{\ast}$ is only employed to bootstrap an initialization of the absolute orientations. 

\subsection{Bootstrapping absolute orientations}\label{sec:init}  
Hartley~\etal~\cite{hartley2011l1} proposed generating a spanning tree by setting the camera with the maximum number of neighbors as the root and recursively adding adjacent cameras without forming a cycle. The reference orientation is fixed at the camera at the root of the spanning tree. The orientations of the rest of the cameras in the tree are computed by propagating away the rectified orientations $R_{uv}^{\ast}$ from the root node along the edges, \ie~$\widetilde R_{v} = R_{uv}^{\ast} \widetilde R_{u}$. 

As discussed before, the noise in the the relative orientation along each edge $R_{uv}^{\ast} \widehat R_{uv}^{-1}$ propagates at the subsequent nodes while computing the initial absolute orientations of the cameras. Therefore, the spanning tree that minimizes the sum of depths of all the nodes (a.k.a. shortest path tree~\cite{tarjan1982sensitivity}) is the best spanning tree for the initialization. Starting with a root node, a shortest path tree could be computed by greedily connecting nodes to each neighboring node in the breadth-first order. The best shortest path tree can be found by applying the same procedure with each one of the nodes as a root node (time complexity $\cO(n^2)$)~\cite{hassin1995minimum}.  However, we employed the procedure just once (time complexity $\cO(n)$) with the root at the node with the maximum number of adjacent nodes (similar to Hartley~\etal~\cite{hartley2011l1}) and observed similar results as with the best spanning tree. The reference orientation of the initialization and the ground-truth is fixed at the root of the tree. This procedure is very fast and it takes only a fraction of a second for a large view-graph with thousands of cameras. We abbreviate this procedure as SPT and it is the default initializer in all of our experiments.  

\subsection{Fine-tuning network}\label{sec:finetune}
The fine-tuning network (FineNet) is again built on a MPNN. It takes the initial absolute orientations $\{\widetilde R_{v}\}$ and the relative orientation measurements $\{\widetilde R_{uv}\}$ as inputs, and predicts the refined absolute orientations $\{R_{v}^{\ast}\}$ as the output. The refined orientations are estimated from the hidden states $h_v^{(T)}$ of the nodes at the last layer of the network, \ie 
\begin{equation} 
R_{v}^{\ast} =  lp_3 \big( h_v^{(T)} \big) \star \widetilde R_{v} 
\label{eq:lp3}
\end{equation} 
where $lp_3$ is a single layer of linear perceptron. 
We initialize the hidden states of the MPNN by the initial orientations, \ie~$h_v^{(0)} = \widetilde R_{v}$. The edge attributes are chosen as the relative discrepancy of the initial and the observed relative orientations, \ie~$e_{uv} = \widetilde R_v^{-1} \widetilde R_{uv} \widetilde R_u$. 
The loss for the fine-tuning network is computed as the weighted sum of edge consistency loss and the rotational distance between the predicted orientation $\widetilde R_{v}$ and the ground-truth orientation $\widetilde R_{v}$, \ie  
\begin{equation}
\cL =  \sum_{\cG \in \cD} \Big( \sum_{\cE_{uv} \in \cE} \cL_{mre}\big( R_{uv}^{\ast}, \widehat R_{uv}\big) + \frac{\beta}{|\cN_v|} \sum_{\cV_v \in \cV}d_{Q}\big( \frac{R_{v}^{\ast}}{\|R_{v}^{\ast}\|}, \widehat R_{v} \big) \Big)  
\end{equation}
where $\cL_{mre}$ is chosen as the quaternion distance~\eqref{eq:quat}. This is a combination of two loss functions chosen according to~\eqref{eq:newloss2}. 
We value consistency of the entire graph (enforced via relative orientations in the first term) over individual accuracy (second term), and so choose $\beta = 0.1$.   



\subsection{Training}
The view-graph cleaning network and the fine-tuning network are trained separately. For each edge $\cE_{uv}$ in the view-graph with observed orientation $\widetilde R_{uv}$, an additional edge $\cE_{vu}$ is included in the view-graph in the opposite direction with  orientation $\widetilde R_{vu} := \widetilde R_{uv}^{-1}$. This will ensure the messages flow in both directions of an edge. In both of the above networks, the parameters are chosen as: the number of time-steps $T=4$, the permutation invariant function $\square$ as the \emph{mean}, and the length of the message $m_v^{(t)}$ and hidden state $h_v^{(t)}$ are $32$. 

\noindent {\bf Network parameters $\Theta$}: The parameters are involved with: (i) 1-D convolutions of $\gamma^{(t)}$ in~\eqref{eq:gamma} and $\phi^{(t)}$ in~\eqref{eq:phi}, and (ii) linear perceptrons $lp_1$, $lp_2$ in~\eqref{eq:lp12} and $lp_3$ in~\eqref{eq:lp3}. With the above hyper-parameters (\ie~time-steps, length of the messages, \etc), the total number of parameters of NeuRoRA becomes ${\approx}49.8$K.  Note that increasing the hyper-parameters could lead to a bigger network (more parameters) and that results a slower performance. A small size network is much faster but is not capable of predicting accurate outputs for larger view-graphs. We have tried different network sizes and found the current network size is a good balance between speed and accuracy. 

\noindent {\bf Architecture setup}: The networks are implemented in \texttt{PyTorch Toolbox}\footnote{\href{https://pytorch-geometric.readthedocs.io/en/latest/}{https://pytorch-geometric.readthedocs.io}} and trained on a GTX 1080 Ti GPU with a learning rate of $0.5 \times 10^{-4}$ and weight decay $10^{-4}$. Each of CleanNet and FineNet are trained for $250$ epochs (takes $\sim 4-6$ hours) to learn the network  parameters. To prevent the networks from over-fitting on the training dataset, we randomly drop $25\%$ of the edges of each view-graph along with observed noisy relative orientations in each epoch.  During testing all the edges were kept active. The parameters that yielded the minimum validation loss were kept for evaluation. All the baselines including the proposed networks were evaluated on an Intel Core i7 CPU. 

\section{Results}\label{sec:results}
Experiments were carried out on synthetic as well as real datasets to demonstrate the true potential of our approach. 

\noindent {\bf Baseline methods} We evaluated NeuRoRA against the following baseline methods (also described in Section~\ref{sec:relative}):  
\begin{itemize}[noitemsep,nolistsep,leftmargin=4.5mm,topsep=0ex,partopsep=0ex]
\item Chatterjee and  Govindu~\cite{chatterjee2017robust}: the latest implementation with the default parameters and cost function in the shared  scripts\footnote{\href{http://www.ee.iisc.ac.in/labs/cvl/research/rotaveraging/}{http://www.ee.iisc.ac.in/labs/cvl/research/rotaveraging/}} were employed. We also employed their evaluation strategy to compare the predicted orientations. 
\item Weiszfeld algorithm~\cite{hartley2011l1}: the algorithm is straightforward but computationally expensive and we only ran it for $50$ iterations  of $\ell_1$ averaging.  
\item Arrigoni~\etal~\cite{arrigoni2018robust}: the authors shared the code with the optimal parameters\footnote{\href{http://www.diegm.uniud.it/fusiello/demo/gmf/}{http://www.diegm.uniud.it/fusiello/demo/gmf/}}. 
\item Wang and Singer~\cite{wang2013exact}: this is employed with a publicly available scripts\footnote{\href{https://github.com/huangqx/map$\_$synchronization}{https://github.com/huangqx/map$\_$synchronization}}. 
\end{itemize}
We also ran the graph cleaning network (CleanNet) followed by bootstrapping initial orientation (using SPT) as a baseline {CleanNet-SPT}, and ran SPT on the noisy graph followed by fine-tuning network (FineNet) as another baseline SPT-FineNet. Note that the proposed network NeuRoRA takes CleanNet-SPT as an initialization and then fine-tunes the initialization in a single step by FineNet. 
{NeuRoRA}-$v2$ is a variation of the proposed method where an initialization from CleanNet-SPT is fine-tuned in two steps of FineNet, \ie~the output of FineNet in the first step is fed as an initialization of FineNet in the second step. 

\noindent {\bf Synthetic dataset} 
We carefully designed a synthetic dataset that closely resembles the real-world datasets. Since the amount of noise in observed relative measurements changes with the sensor type (\eg~camera device), the structure of the connections in the view-graphs and the outlier ratios are varied with the scene (Figure~\ref{fig:statistics}).  
A single view-graph was  generated as follows: (1) the number of cameras were sampled in the range  $250-1000$ and their orientations were generated randomly on a horizontal plane (yaw only), (2) pairwise edges and corresponding relative orientations were randomly introduced between the cameras that amounted to $(10-30)\%$ of  all possible pairs, (3) the relative orientations were then corrupted by a noise with a std $\sigma$ where $\sigma$ is chosen uniformly in the range $(5\degree-30\degree)$ once for the entire view-graph, and the directions are chosen randomly on the vertical plane (to emulate realistic distributions~\ref{fig:statistics}), and (4) the relative orientations were further corrupted by $(0 - 30)\%$ of outliers with random orientations. 
Our synthetic dataset consisted of $1200$ sparse view-graphs. The dataset was divided into training ($80\%$), validation ($10\%$), and testing ($10\%$). 

The results are furnished in Table~\ref{tab:results}. The average angular error on all the view-graphs in the dataset is displayed. The proposed method NeuRoRA performs remarkably well compared to the baselines in terms of accuracy and speed. NeuRoRA-$v2$ further improves the results. Overall, Chatterjee~\cite{chatterjee2017robust} performs well but the performance does not improve with a better initialization. Unlike Wang~\cite{wang2013exact}, Weiszfeld~\cite{hartley2011l1} improves the performance with a better initialization given by CleanNet-SPT, but, it can not improve the solution further given an even better initialization by NeuRoRA. 
Notice that the proposed NeuRoRA is three orders of magnitude faster with a GPU than the baseline methods.  
\begin{table}[H]\setlength{\tabcolsep}{4.8pt}
\caption{Results of Rotation averaging on a test synthetic dataset. The average angular error on all the view-graphs in our dataset is displayed. The proposed method NeuRoRA is remarkably faster than the baselines while producing better results. 
There is no GPU implementations of \cite{chatterjee2017robust,hartley2011l1,arrigoni2018robust,wang2013exact} available, thus the runtime comparisons on cuda are excluded. Note that NeuRoRA takes only $\bm{0.0016s}$ on average on a GPU.} 
\centering  \tiny   
\begin{tabular}{@{\hspace{0.1em}}p{2.6cm}@{\hspace{0.5em}}c@{\hspace{0.5em}}c@{\hspace{0.5em}}c@{\hspace{0.5em}}c|p{2.45cm}@{\hspace{0.5em}}c@{\hspace{0.5em}}c@{\hspace{0.5em}}c@{\hspace{0.5em}}c@{\hspace{0.5em}}}
\toprule
{\bf Baseline Methods} & mn & md & \multicolumn{2}{c}{cpu} &  & mn & md & \multicolumn{2}{c}{cpu} \\ 
\midrule 
{Chatterjee}~\cite{chatterjee2017robust}  &   $2.17\degree$ & $1.25\degree$ & $5.38$s &$(~~~~1\times)$ & 
{Arrigoni}~\cite{arrigoni2018robust} &  $2.92\degree$ & $1.42\degree$ & $8.20s$ & $(0.65\times)$  \\
{Weiszfeld}~\cite{hartley2011l1} &  $3.35\degree$ & $1.02\degree$ & $50.92s$ & $ ({0.11\times})$  & 
{Wang}~\cite{wang2013exact}  &   $2.77\degree$ & $1.40\degree$ & $9.75$s &$(0.55\times)$  \\ \midrule 
{\bf Proposed Methods} & \\ 
\midrule 
CleanNet-SPT~+~\cite{chatterjee2017robust}  &   $2.11\degree$ & $1.26\degree$ &  $5.41s$ & $(0.99\times)$ &  
{NeuRoRA}  & $\bm{1.45\degree}$ & $\bm{0.74\degree}$ & $\bm{0.21s}$ & $(~~\textcolor{black}{\bm{24\times}})$ \\
CleanNet-SPT~+~\cite{hartley2011l1} &  $1.74\degree$ & $1.01\degree$  & $50.36s$ & $({0.11\times}$) &
NeuRoRA-$v2$  & $\bm{1.30\degree}$ & $\bm{0.68\degree}$  & $\bm{0.30s}$ & $(~~{\bm{18\times}})$ \\ \midrule 
\bf Other Methods & \\ \midrule  
{CleanNet-SPT} & $2.93\degree$ & $1.47\degree$ & $\bm{0.11s}$ & $(~~{\bm{47\times}})$ &  
{SPT-FineNet} & $3.00\degree$ & $1.57\degree$ &    $\bm{0.11s}$ & $(~~{\bm{47\times}})$ \\  
{CleanNet-SPT}~+~\cite{wang2013exact} & $2.77\degree$ & $1.40\degree$ & ${9.86s}$ & $({{0.53\times}})$ &  
{SPT-FineNet}~+~\cite{wang2013exact} & $2.78\degree$ & $1.40\degree$ &    ${9.86s}$ & $({{0.53\times}})$ \\  
SPT-FineNet~+~\cite{chatterjee2017robust}  &   $2.12\degree$ & $1.26\degree$ & ~$5.41s$ & $(0.99\times)$ & 
SPT-FineNet~+~\cite{hartley2011l1} &  $1.78\degree$ & $1.01\degree$ & $50.36s$ & $({0.11\times})$  \\ 
{NeuRoRA~+~\cite{chatterjee2017robust}}  & $2.11\degree$ & $1.26\degree$  & ~${5.51s}$ & $(0.97\times)$  &  
{NeuRoRA~+~\cite{hartley2011l1}}  & ${1.73\degree}$ & ${1.01\degree}$  & ${50.46s}$ & $({0.10\times})$  \\ \midrule 
\multicolumn{10}{c}{mn: mean of the angular error; md: median of the angular error; cpu: the average runtime } \\
\multicolumn{10}{c}{of the method; MethodA~+~MethodB: MethodB is initialized by the solution of MethodA} \\ 
 \bottomrule
\end{tabular} 
\label{tab:results} 
\end{table} 

\noindent {\bf Real dataset} 
We summarize the real datasets and display in Table~\ref{tab:real_results}.  There are a total of $19$ publicly available view-graphs with observed noisy relative orientations and the ground-truth absolute orientations. The ground-truth orientations were obtained by applying  incremental bundle adjustment~\cite{agarwal2010bundle} on the view-graphs. The {TNotreDame} dataset is shared by Chatterjee~\etal~\cite{chatterjee2017robust}\footnote{\href{http://www.ee.iisc.ac.in/labs/cvl/research/rotaveraging/}{http://www.ee.iisc.ac.in/labs/cvl/research/rotaveraging/}}. The {Artsquad} and {SanFrancisco} datasets are provided by DISCO~\cite{crandall2011discrete}\footnote{\href{http://vision.soic.indiana.edu/projects/disco/}{http://vision.soic.indiana.edu/projects/disco/}}. The rest of the view-graphs are publicly shared by 1DSFM~\cite{wilson2014robust}\footnote{\href{http://www.cs.cornell.edu/projects/1dsfm/}{http://www.cs.cornell.edu/projects/1dsfm/}}. 
The ground-truth orientations are available for some of those cameras (indicated in parenthesis) and the training, validation and testing are  performed only on those cameras. 

Due to limited availability of real datasets for training, we employed network parameters pre-trained on the above synthetic dataset and further fine-tuned on the real datasets in round-robin fashion (leave one out).  Such evaluation protocol is employed because we did not want to divide the sequences into training and testing sequences that might favor one particular method.
The finetuning is done for each round of the round-robin using the real-data apart from the held-out test sequence.
Overall, the proposed NeuRoRA outperformed the baselines for this task in terms of accuracy and efficiency (Table~\ref{tab:real_results}). The {Artsquad} and {SanFrancisco} datasets have different orientation patterns as shared from a different source~\cite{crandall2011discrete}. In particular {SanFrancisco} dataset is captured along a road which is significantly different from others. Thus, the performance of NeuRoRA falls short to  Chatterjee~\cite{chatterjee2017robust} and Wang~\cite{wang2013exact} only on those two sequences, 
but, is still better than Weiszfeld~\cite{hartley2011l1} and Arrigoni~\cite{arrigoni2018robust}. 
Nonetheless, the proposed NeuRoRA is much faster than others. 

\begin{table}[H] \setlength{\tabcolsep}{1.1pt}
\caption{Results of MRA on real datasets. The proposed method NeuRoRA is much faster than the baselines while producing overall similar or better results. 
The number of cameras, for which ground-truths are available, is shown within parenthesis.} 
\centering  \tiny   
\begin{tabular}{@{\hspace{0.1em}} p{1.6cm}@{\hspace{-0.85em}}r@{\hspace{0.25em}}r|r@{\hspace{0.75em}}r@{\hspace{0.25em}}r|rr@{\hspace{0.25em}}r|rr@{\hspace{0.25em}}r|rr@{\hspace{0.25em}}r|rr@{\hspace{0.25em}}r}
\toprule 
 \multicolumn{3}{c|}{Datasets}&\multicolumn{3}{c|}{{ Chatterjee}~\cite{chatterjee2017robust}}&\multicolumn{3}{c|}{{ Weiszfeld}~\cite{hartley2011l1}} &\multicolumn{3}{c|}{{Arrigoni}~\cite{arrigoni2018robust}}&\multicolumn{3}{c|}{{Wang}~\cite{wang2013exact}}  & \multicolumn{3}{c}{NeuRoRA}\\ \midrule 
Name & \texttt{\#}cameras & \texttt{\#}edges & mn & md & cpu &   mn & md & cpu   &  mn & md & cpu &   mn & md & cpu   &  mn & md & cpu \\ 
Alamo & $627 (577)$ & $49.5\%$ & {$\bm{4.2}$} & $\bm{1.1}$  & $ 20.5s $ & $ 4.9 $ & $ 1.4 $  & $ 84.0s
$& $ 6.2 $ & $ 1.6 $  & $ 2.7s $ & $ 5.3 $ & $ 1.4 $ & $ 20.6s
 $ & $ 4.9 $ & $ 1.2 $ & $ \bm{2.2s} $ \\
 
EllisIsland  & $ 247 (227) $ & $ 66.8\%	$ & $  2.8 $ & $  \bm{ 0.5} $  & $ 2.5s $ & $ 4.4 $ & $ 1.0 $ & $ 8.9s 
$ & $ 3.9 $ & $ 1.2 $  & $ \bm{0.2s} $ & $ 3.6 $ & $ 1.1 $ & $ 2.6s
 $ & $ \bm{2.6} $ & $ 0.6 $ & $ {0.4s} $ \\ 
 
GendrmMarkt & $ 742(677)$ & $ 17.5\%  $ & $ \textcolor{red}{37.6} $ & $ {7.7} $  & $ 11.1s $ & $ \textcolor{red}{29.4} $ & $ {9.6} $  & $ 53.7s
 $ & $ \textcolor{red}{41.6} $ & $ 13.3 $  & $ 8.9s $ & $ \textcolor{red}{32.6} $ & $ 6.1 $ & $ 12.5s
 $& $ \bm{4.5} $ & $ \bm{2.9} $  & $ \bm{0.5s} $ \\
 
MadridMetrop & $ 394(341)$ & $ 30.7\% $ & $ 6.9 $ & $ 1.2 $  & $ 3.2s $ & $ 7.5 $ & $ 2.7 $  & $ 14.5s 
$& $ 6.0 $ & $ 1.7 $  & $ 0.9s $ & $ 5.0 $ & $ 1.4 $ & $ 3.6s
 $ & $ \bm{2.5} $ & $ \bm{1.1} $ &  $ \bm{0.2s } $\\ 
 
MontrealNotre  & $ 474 (450) $ & $  46.8\% $ & $ 1.5 $ & $ \bm{0.5} $ &  $ 9.1s $ & $ 2.1 $ & $ 0.7 $ & $ 41.5s
 $& $ 4.8 $ & $ 0.9 $  & $ 2.9s $ & $ 2.0 $ & $ 0.8 $ & $ 10.1s
 $ & $ \bm{1.2} $ & $ 0.6 $ & $ \bm{1.0s } $ \\ 
 
NYCLibrary & $ 376 (332) $ & $  29.3\% $ & $ 3.0 $ & $ 1.3 $  & $ 4.8s $ & $ 3.8 $ & $ 2.1 $ & $ 14.4s
 $& $ 3.9 $ & $ 1.5 $  & $ 1.4s $ & $ 2.9 $ & $ 1.4 $ & $ 3.2s
 $ & $ \bm{1.9} $ & $ \bm{1.1} $  & $ \bm{0.2s } $ \\
 
NotreDame & $ 553 (553) $ & $  68.1\%$ & $ 3.5 $ & $ \bm{0.6} $  & $ 23.3s $ & $ 4.7 $ & $ 0.8 $  & $ 80.8s
$ & $ 3.9 $ & $ 1.0 $  & $ 4.2s $ & $ 3.5 $ & $ 0.9 $ & $ 19.5s
 $& $ \bm{1.6} $ & $ \bm{0.6} $ &  $ \bm{2.0s} $\\
 
PiazzaDelP & $ 354 (338) $ & $ 39.5\% $ & $ 4.0 $ & $ 0.8 $ &  $ 3.3s $ & $ 4.8 $ & $ 1.3 $  & $ 16.7s
 $& $ 10.8 $ & $ 1.2 $  & $ 0.6s $ & $ 6.2 $ & $ 1.1 $ & $ 3.6s
 $ & $ \bm{3.0} $ & $ \bm{0.7} $  & $ \bm{0.4s} $\\ 
 
Piccadilly & $ 2508 (2152) $ & $ 10.2\%	$ & $ 6.9 $ & $ 2.9 $  & $ \textcolor{red}{449.0s} $ & $ \textcolor{red}{26.4} $ & $ 7.5 $ &  ${\sim}\textcolor{red}{20m} 
$& $ \textcolor{red}{22.0} $ & $ 9.7 $  & $ 43.7s $ & $ 10.1 $ & $ 3.9 $ & $ 118.1s
 $ & $ \bm{4.7} $ & $ \bm{1.9} $  & $ \bm{5.9s} $\\
 
RomanForum  & $ 1134 (1084)$ & $  10.9\% 	$ & $ 3.1 $ & $ 1.5 $  & $ 20.2s $ & $ 4.8 $ & $ 1.8 $ & $ 115.0s
$& $ 13.2 $ & $ 8.2 $  & $ 16.8s $ & $ 4.6 $ & $ 3.5 $ & $ 19.6s
 $ & $ \bm{2.3} $ & $ \bm{1.3} $ & $ \bm{1.3s} $\\
 
TowerLondon  & $ 508 (472) $ & $ 18.5\% $ & $ 3.9 $ & $ 2.4 $  & $ 1.9s $ & $ 4.7 $ & $ 2.9 $  & $ 17.1s
$ & $ 4.6 $ & $ 1.8 $  & $ 3.9s $ & $ 2.9 $ & $ 1.5 $ & $ 3.6s
 $& $ \bm{2.6} $ & $ \bm{1.4} $ &  $ \bm{0.3s} $\\
 
Trafalgar  & $ 5433 (5058) $ & $ 4.6\%	$ & $ \bm{3.5} $ & $ \bm{2.0} $  & $ \textcolor{red}{858.4s} $ & $ {15.6} $ & $ \textcolor{red}{11.3} $  & ${\sim}\textcolor{red}{92m} 
$& $ \textcolor{red}{48.6} $ & $ 13.2 $  & $ \textcolor{red}{167.4s} $ & $ 17.2 $ & $ 16.0 $ & $ \textcolor{red}{319.2s}
 $ & $ 5.3 $ & $ {2.2} $ &  $ \bm{15.5s} $ \\
 
UnionSquare  & $ 930 (789) $ & $5.9\%$ & $ {9.3} $ & $ 3.9 $  & $ 6.8s $ & $ \textcolor{red}{40.9} $ & $ {10.3} $  & $ 42.8s
$& $ 9.2 $ & $ 4.4 $  & $ 12.1s $ & $ 6.8 $ & $ 3.2 $ & $ 4.1s
 $ & $ \bm{5.9} $ & $ \bm{2.0} $  & $ \bm{0.6s }$\\ 
 
ViennaCath  & $ 918 (836) $ & $  24.6\% $ & $ {8.2} $ & $ 1.2 $  & $ 48.1s $ & $ {11.7} $ & $ 1.9 $  & $ \textcolor{red}{158.3s} 
$ & $ 19.3 $ & $ 2.39 $  & $ 6.0s $ & $ 10.1 $ & $ 1.8 $ & $ 25.7s
 $& $ \bm{3.9} $ & $ \bm{1.5} $ & $ \bm{2.1s }$\\
 
Yorkminster  & $ 458 (437) $ & $ 26.5\% 	$ & $ 3.5 $ & $ 1.6 $  & $ 4.0s $ & $ 5.7 $ & $ 2.0 $  & $ 32.0s 
$& $ 4.5 $ & $ 1.6 $  & $ 2.5s $ & $ 3.5 $ & $ 1.3 $ & $ 4.9s
 $ & $ \bm{2.5} $ & $ \bm{0.9} $ &  $ \bm{ 0.4s} $ \\ 
 
Acropolis  & $ 463 (463) $ & $ 10.7\% $ & $ 1.1 $ & $ 0.7 $ & $ 1.5s $ & $ \bm{0.6} $ & $ \bm{0.3} $  & $ 15.0s 
$ & $ 2.7 $ & $ 1.7 $  & $ 3.2s $ & $ 2.4 $ & $ 1.6 $ & $ 1.7s
 $ & $ 0.8 $ & $ 0.5 $  & $\bm{ 0.2s}$\\ 
 
ArtsQuad  & $ 5530 (4978) $ & $1.4\%$ & $ \bm{4.8} $ & $ {3.5} $ & $ 116.1s $ & $ \textcolor{red}{34.4} $ & $ \textcolor{red}{23.1} $  & ${\sim}\textcolor{red}{32m} 
$& $ \textcolor{red}{35.2} $ & $ 15.8 $  & $ \textcolor{red}{189.4s} $ & $ 6.0 $ & $ \bm{3.2} $ & $ 73.9s
 $ & $ \textcolor{red}{27.5} $ & $ 7.3 $  & $ \bm{5.0s}$ \\
 
SanFran  & $ 7866 (7866) $ & $ 0.3\% $ & $ \bm{3.6} $ & $ \bm{3.4} $  & $ 15.2s $ & $ {18.8}$ & $ {16.4} $  & ${\sim}\textcolor{red}{22m} 
$& $ \textcolor{red}{66.8} $ & $ \textcolor{red}{43.9} $  & $ \textcolor{red}{354.7s} $ & $ \textcolor{red}{89.2} $ & $ \textcolor{red}{75.5} $ & $ 27.2s
 $ & $ {17.6} $ & $ {12.6} $ & $ \bm{2.6s}$\\
 
TNotreDame  & $ 715 (715) $ & $ 25.3\%$ & $ \bm{1.0} $ & $ \bm{0.4} $  & $ 10.6s $ & $ 1.4 $ & $ 0.6 $ &  $ 72.5s 
$& $ 2.4 $ & $ 0.9 $  & $ 5.7s $ & $ 1.7 $ & $ 0.8 $ & $ 14.8s
 $ & $ 1.7 $ & $ 0.7 $ & $ \bm{1.4s} $\\ \midrule 
\multicolumn{18}{c}{mn: mean of the angular error (in deg); md: median of the angular error (in deg); cpu: the runtime of} \\
\multicolumn{18}{c}{ the method on a cpu ($s$: in sec, $m$: in minute); entries with $>20\deg$ or $>120s$ are marked in \textcolor{red}{red}.} \\ 
\bottomrule
\end{tabular} 
\label{tab:real_results} 
\end{table} 
\noindent {\bf Robustness Check} 
In this experiment we study the generalization capability of NeuRoRA. To check the individual effects of different sensor settings, we generate a number of synthetic dataset varying {\bf (i)} \texttt{\#}cameras {\bf (ii)} \texttt{\#}edges, {\bf (iii)} amount of noise and outliers, and {\bf (iv)} planar/random camera motion. NeuRoRA is then trained on one of such datasets and evaluated on the others. Each dataset consists of $1000$ view-graphs (large ones contain only $100$). Results are furnished in Table~\ref{tab:robustness}. Notice that NeuRoRa  generalizes well across dataset changes except when the network is trained on planar cameras and tested on random. We therefore advice to use two separate networks for planar and non-planner scenes. 
Notice that {Chatterjee}~\cite{chatterjee2017robust} demands a large memory for large view-graphs and failed to execute on a system of 64gb of RAM. We tested our method on view-graphs upto $25$K vertices and $24$M edges on the same system. 

\begin{table}[H] \setlength{\tabcolsep}{3.8pt}
\caption{Robustness check: results of Rotation averaging on multiple datasets where NeuRoRA is trained on one and evaluated on the other datasets.} 
\centering  \tiny   
\begin{tabular}{@{\hspace{0.1em}}p{1.55cm}@{\hspace{0.75em}}r@{\hspace{0.75em}}r@{\hspace{0.75em}}c@{\hspace{0.75em}}c@{\hspace{0.2em}}r@{\hspace{0.75em}}r@{\hspace{0.75em}}r@{\hspace{0.75em}}r@{\hspace{0.75em}}|r@{\hspace{0.5em}}r@{\hspace{0.2em}}r@{\hspace{0.5em}}|r@{\hspace{0.5em}}r@{\hspace{0.5em}}r@{\hspace{0.5em}}}
\toprule
 & \multicolumn{4}{c}{Training datasets}  & \multicolumn{4}{c}{Evaluation datasets}  & \multicolumn{3}{c}{NeuRoRA} & \multicolumn{3}{c}{{ Chatterjee}~\cite{chatterjee2017robust}}  \\ 
 \cmidrule[0.06em](lr){2-5}\cmidrule[0.06em](lr){6-9} \cmidrule[0.06em](lr){10-12} \cmidrule[0.06em](lr){13-15} 
{\bf Robustness } & $|\cV|$ & $|\cE|$~~ & E${\&}$O~~ & P & $|\cV|$ & $|\cE|$~~ & E${\&}$O~~ & P & mn~ & md~ & cpu & mn~ & md~ & cpu\\ \midrule 
\texttt{\#}cameras  & $1000$ & $25.0\%$ & $30^\circ\&10\%$ & \checkmark & $\bm{250}$ & $25.0\%$ & $30^\circ\&10\%$ & \checkmark & $\bm{1.1\degree}$ & $\bm{0.9\degree}$ & $\bm{0.1s}$ & $1.8\degree$ & $1.7\degree$ & ${0.3s}$\\ 
($|\cV|$)  & $\bm{250}$ & $25.0\%$ & $30^\circ\&10\%$ & \checkmark & $\bm{5000}$ & $\bm{2.5\%}$ & $30^\circ\&10\%$ & \checkmark & $\bm{1.1\degree}$ & $\bm{1.0\degree}$ & $\bm{4.9s}$ & $1.4\degree$ & $1.3\degree$ & ${\sim}\textcolor{red}{12m}$\\ 
  & $\bm{250}$ & $25.0\%$ & $30^\circ\&10\%$ & \checkmark & $\bm{10000}$ & $\bm{2.5\%}$ & $30^\circ\&10\%$ & \checkmark & $\bm{0.7\degree}$ & $\bm{0.6\degree}$ & $\bm{18.7s}$ & \multicolumn{3}{c}{\textcolor{red}{Out of memory}} \\ 
 & $\bm{250}$ & $25.0\%$ & $30^\circ\&10\%$ & \checkmark & $\bm{25000}$ & $\bm{2.5\%}$ & $30^\circ\&10\%$ & \checkmark & $\bm{0.6\degree}$ & $\bm{0.5\degree}$ & $\bm{142.6s}$ & \multicolumn{3}{c}{\textcolor{red}{Out of memory}}\\   
  \midrule

\texttt{\#}edges  & $1000$ & $25.0\%$ & $30^\circ\&10\%$ & \checkmark & $1000$ & $\bm{2.5\%}$ & $30^\circ\&10\%$ & \checkmark & $\bm{2.4\degree}$ & $\bm{2.1\degree}$ & $\bm{0.1s}$ & $3.0\degree$ & $2.8\degree$ & ${3.3s}$ \\
       ($|\cE|$)        & $1000$ & $\bm{2.5\%}$ & $30^\circ\&10\%$ & \checkmark & $1000$ & $25.0\%$ & $30^\circ\&10\%$ & \checkmark & $\bm{0.5\degree}$ & $\bm{0.4\degree}$ & $\bm{2.5s}$ & $0.9\degree$ & $0.8\degree$ & ${43.2s}$ \\ \midrule 
noise~\texttt{\&}~outliers  & $1000$ & $25.0\%$ & $30^\circ\&10\%$ & \checkmark & $1000$ & $25.0\%$ & $\bm{10^\circ\&5\%}$ & \checkmark & ${0.4\degree}$ & $\bm{0.3\degree}$ & $\bm{2.5s}$ & $\bf{0.3\degree}$ & $\bf{0.3\degree}$ & ${31.6s}$ \\ 
       (E${\&}$O)        & $1000$ & $25.0\%$ & $\bm{10^\circ\&~~5\%}$ & \checkmark & $1000$ & $25.0\%$ & $30^\circ\&10\%$ & \checkmark & $\bm{0.6\degree}$ & $\bm{0.5\degree}$ & $\bm{2.5s}$ & ${0.9\degree}$ & $0.8\degree$ & ${43.2s}$ \\ \midrule 
planar  & $1000$ & $25.0\%$ & $30^\circ\&10\%$ & \checkmark & $1000$ & $25.0\%$ & $30^\circ\&10\%$ & \ding{55} & ${2.2\degree}$ & ${1.6\degree}$ & $\bm{2.5s}$ & $\bm{0.9\degree}$ & $\bm{0.8\degree}$ & ${26.3s}$ \\ 
      (P)         & $1000$ & $25.0\%$ & $30^\circ\&10\%$ & \ding{55} & $1000$ & $25.0\%$ & $30^\circ\&10\%$ & \checkmark & $\bm{0.9\degree}$ & $\bm{0.7\degree}$ & $\bm{2.5s}$ & $\bm{0.9\degree}$ & ${0.8\degree}$ & ${26.3s}$ \\ 
 \midrule
 \multicolumn{15}{c}{E: noise with a std chosen uniformly in the range $(0\degree-\text{E}\degree)$; O: percentage of outliers; P: flag for planar cameras} \\
 \bottomrule
\end{tabular} 
\label{tab:robustness} 
\end{table} 

\section{Discussion}
We have proposed a graph-based neural network for absolute orientation regression of a number of cameras from their observed relative orientations. 
The proposed network is exceptionally faster than the strong optimization-based baselines while producing better results on most datasets. 
The outstanding performance of the current work and the relevant neural networks for test-time optimization leads to the following question:  \emph{``can we then replace all the optimizations in robotics / computer vision by a suitable neural network-based regression?''} The answer is obviously \emph{No}. For instance, if an optimization at test-time requires solving a simpler convex cost with a few parameters to optimize, a naive gradient descent will find the globally optimal parameters, while a network-based regression would only estimate sub-optimal parameters. 
To date, neural nets have been proven to be consistently better at solving pattern recognition problems than solving a constraint optimization problems. A few neural network-based solutions are proposed recently that can exploit the patterns in the data while solving a test-time optimization. Therefore the current work also opens up many questions related to the right tool for a specific application.  
\section*{Acknowledgement}
We gratefully acknowledge the support of the Australian Research Council through the Centre of Excellence for Robotic Vision, CE140100016 and the Australian Research Council Discovery Project DP200101675. 


\appendix

\section{Results on ScanNet datasets}
The dataset~\cite{dai2017scannet} consists of $2.5$M images from $1.5$K indoor scenes. The images are collected by capturing videos with a tablet.  An RGBD SLAM system (BundleFusion~\cite{dai2017bundlefusion}) is then applied to get the ground-truth camera poses (along with the absolute orientations). The dataset however does not provide the relative poses of the images. We employed a publicly available toolbox VisualSFM~\cite{wu2011visualsfm} to obtain the relative orientations of consecutive images. Note that only $50$ neighboring images for each image are used for the comparison. For each sequence only the largest connected component was considered to form our dataset. Note that the pair-wise comparison is slow, and so far we have collected $480$ sequences (out of $1500$) of which $40\%$ were used for training, $10\%$ for validation and remaining $50\%$ were used for testing. 
The average number of \texttt{\#}cameras is $1401$ and average number of edges is \texttt{\#}edges is $5.8\%$.  

The results are displayed in Figure~\ref{tab:results}.  Note that these sequences are very tough sequences as these are very sparsely connected forming  chains.  The proposed CleanNet does a very good job cleaning the network and FineNet fine-tunes it. The combined network NeuRoRa produces best results compared to the baselines. Although the performance gap w.r.t. {Chatterjee}~\cite{chatterjee2017robust} is small, NeuRoRa is much faster. We will release the entire dataset (once built) public to foster research in this direction.

\section{The angle and axes of observed relative orientations}

More examples of samples of Figure~2 in the main paper are plotted in Figure~\ref{fig:statistics}. 

\begin{figure}
\centering \scriptsize   
\begin{tabular}{l@{\hspace{0.1em}}c@{\hspace{0.1em}}c@{\hspace{1em}}c@{\hspace{0.1em}}c@{\hspace{1em}}c@{\hspace{0.1em}}c@{\hspace{1em}}c@{\hspace{0.1em}}c} 

\begin{picture}(1,20)\put(-5, -12){\rotatebox{90}{~~~~~~~~Relative~~~}}\end{picture} & 
{\includegraphics[width=0.115\textwidth, height=1.44cm]{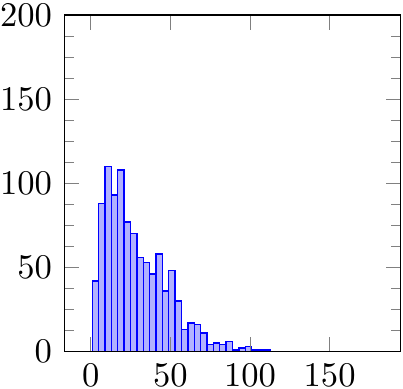}} & 
{\includegraphics[width=0.115\textwidth, height=1.6cm]{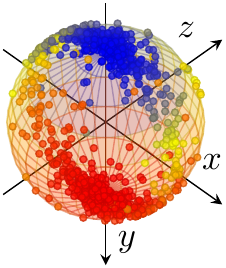}} & 
{\includegraphics[width=0.115\textwidth, height=1.44cm]{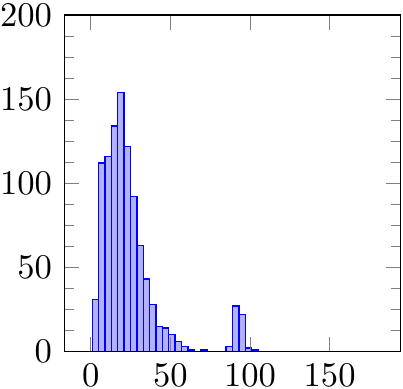}} & 
{\includegraphics[width=0.115\textwidth, height=1.6cm]{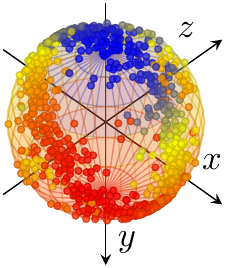}} & 
{\includegraphics[width=0.115\textwidth, height=1.44cm]{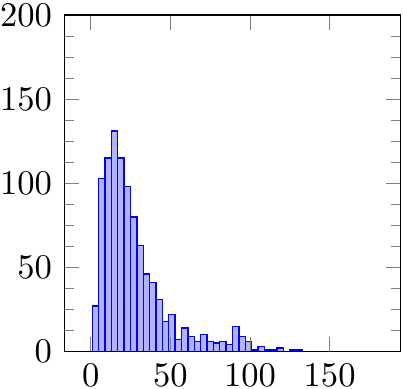}} & 
{\includegraphics[width=0.115\textwidth, height=1.6cm]{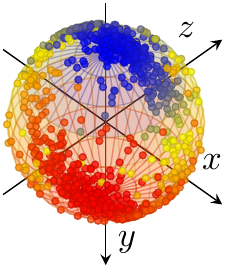}} & 
{\includegraphics[width=0.115\textwidth, height=1.44cm]{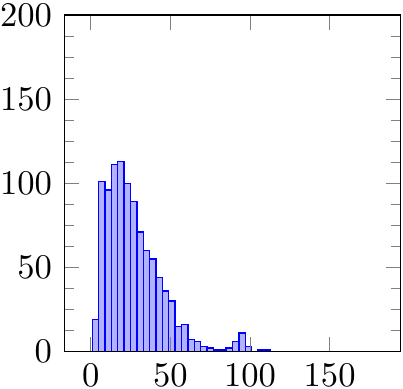}} & 
{\includegraphics[width=0.115\textwidth, height=1.6cm]{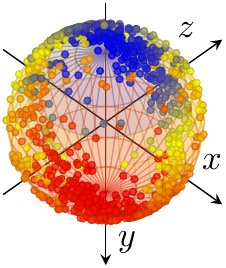}} \\

\begin{picture}(1,20)\put(-5, -12){\rotatebox{90}{~~~~~~~~~Noise~~~}}\end{picture} & 
{\includegraphics[width=0.115\textwidth, height=1.44cm]{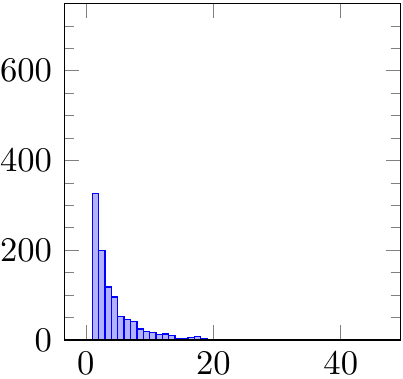}} & 
{\includegraphics[width=0.115\textwidth, height=1.6cm]{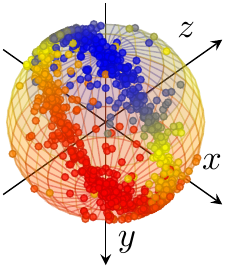}} & 
{\includegraphics[width=0.115\textwidth, height=1.44cm]{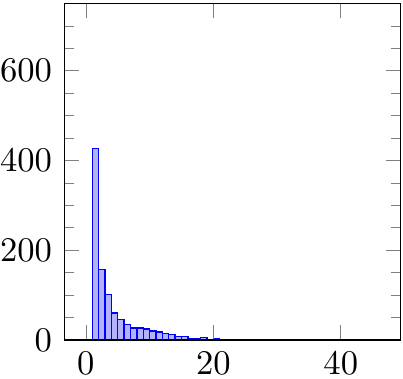}} & 
{\includegraphics[width=0.115\textwidth, height=1.6cm]{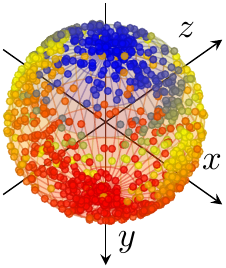}} & 
{\includegraphics[width=0.115\textwidth, height=1.44cm]{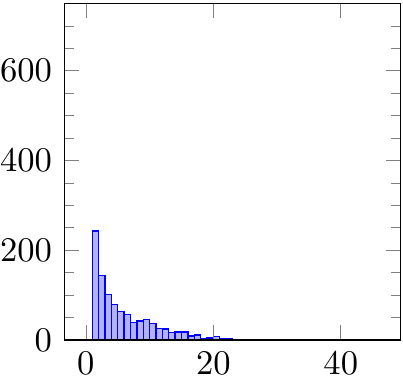}} & 
{\includegraphics[width=0.115\textwidth, height=1.6cm]{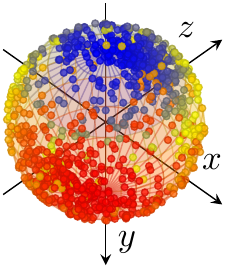}} & 
{\includegraphics[width=0.115\textwidth, height=1.44cm]{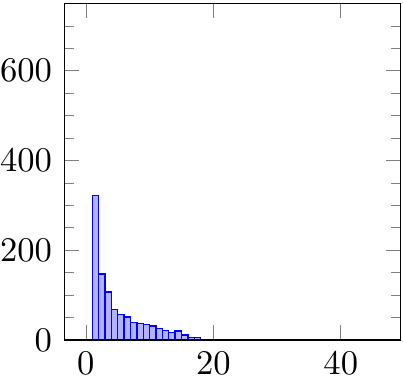}} & 
{\includegraphics[width=0.115\textwidth, height=1.6cm]{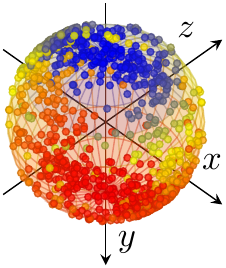}} \\
& \multicolumn{2}{c}{(a) Alamo}  &  \multicolumn{2}{c}{(b) Ellis Island}  &  \multicolumn{2}{c}{(c) Gendermen Market}  &  \multicolumn{2}{c}{(d) Madrid Metropolis}  \\  ~\\
\begin{picture}(1,20)\put(-5, -12){\rotatebox{90}{~~~~~~~~Relative~~~}}\end{picture} & 
{\includegraphics[width=0.115\textwidth, height=1.44cm]{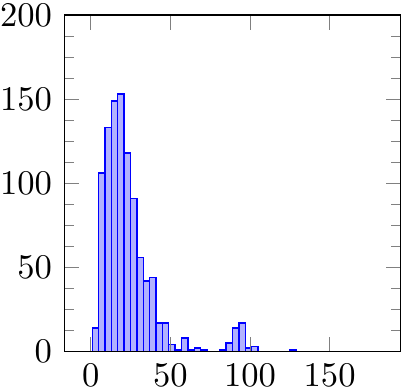}} & 
{\includegraphics[width=0.115\textwidth, height=1.6cm]{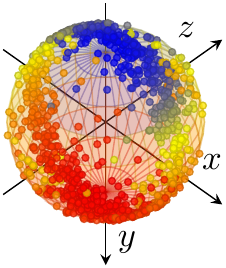}} & 
{\includegraphics[width=0.115\textwidth, height=1.44cm]{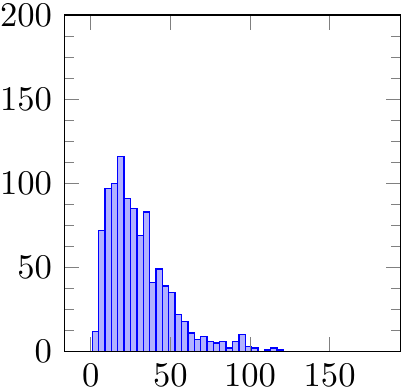}} & 
{\includegraphics[width=0.115\textwidth, height=1.6cm]{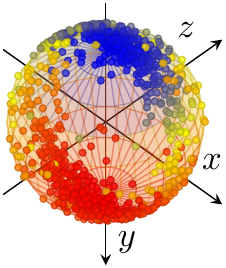}} & 
{\includegraphics[width=0.115\textwidth, height=1.44cm]{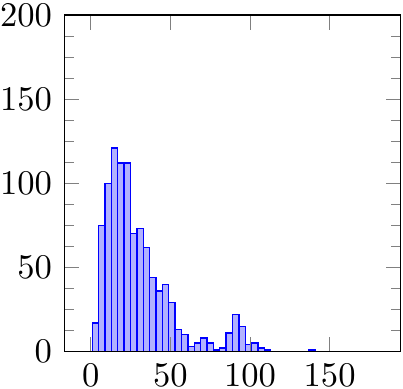}} & 
{\includegraphics[width=0.115\textwidth, height=1.6cm]{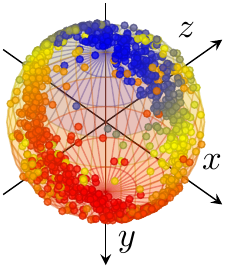}} & 
{\includegraphics[width=0.115\textwidth, height=1.44cm]{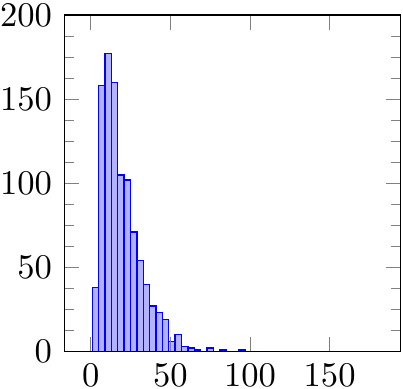}} & 
{\includegraphics[width=0.115\textwidth, height=1.6cm]{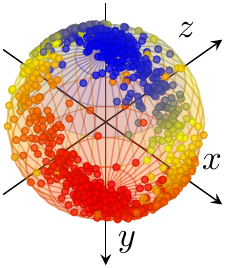}} \\

\begin{picture}(1,20)\put(-5, -12){\rotatebox{90}{~~~~~~~~~Noise~~~}}\end{picture} & 
{\includegraphics[width=0.115\textwidth, height=1.44cm]{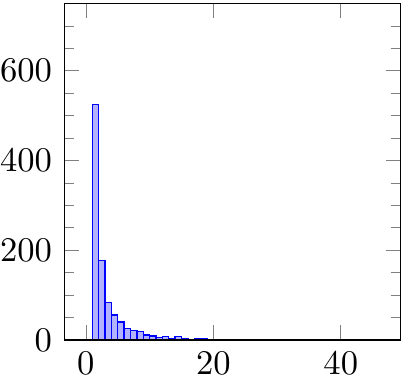}} & 
{\includegraphics[width=0.115\textwidth, height=1.6cm]{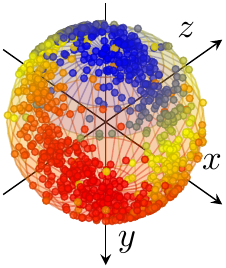}} & 
{\includegraphics[width=0.115\textwidth, height=1.44cm]{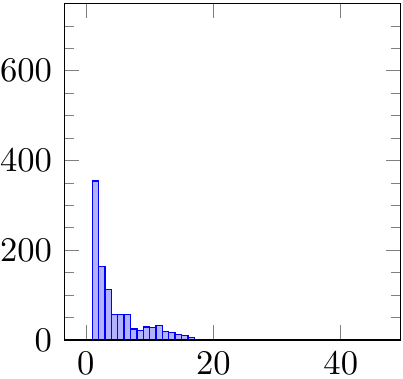}} & 
{\includegraphics[width=0.115\textwidth, height=1.6cm]{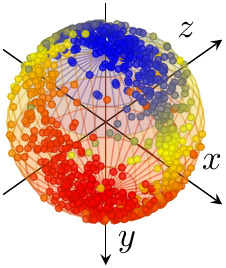}} & 
{\includegraphics[width=0.115\textwidth, height=1.44cm]{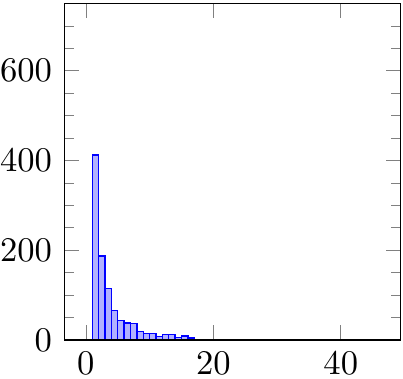}} & 
{\includegraphics[width=0.115\textwidth, height=1.6cm]{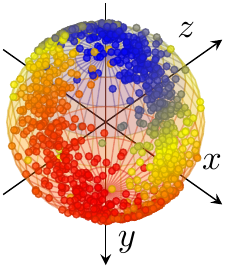}} & 
{\includegraphics[width=0.115\textwidth, height=1.44cm]{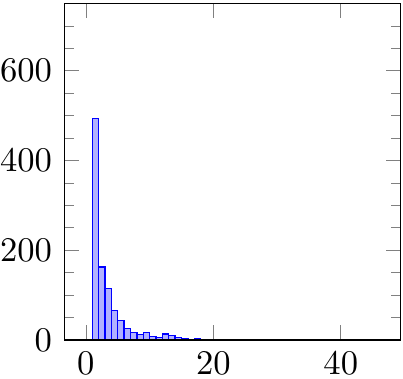}} & 
{\includegraphics[width=0.115\textwidth, height=1.6cm]{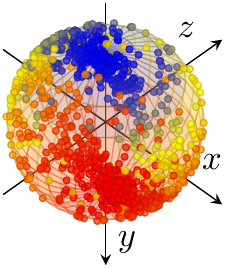}} \\
& \multicolumn{2}{c}{(e) MontrealNotreDam}  &  \multicolumn{2}{c}{(f) NYC Library}  &  \multicolumn{2}{c}{(g) Notre Dame}  &  \multicolumn{2}{c}{(h) Piazza Del Popolo}  \\  ~\\
\begin{picture}(1,20)\put(-5, -12){\rotatebox{90}{~~~~~~~~Relative~~~}}\end{picture} & 
{\includegraphics[width=0.115\textwidth, height=1.44cm]{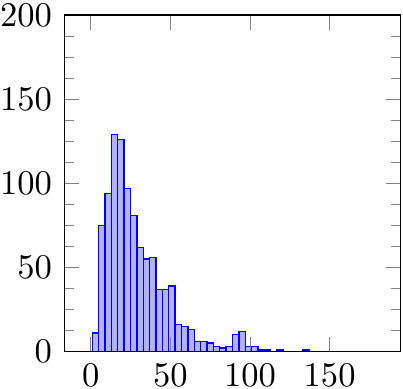}} & 
{\includegraphics[width=0.115\textwidth, height=1.6cm]{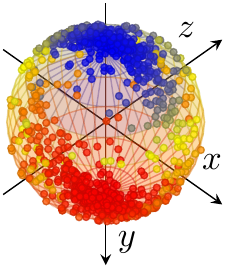}} & 
{\includegraphics[width=0.115\textwidth, height=1.44cm]{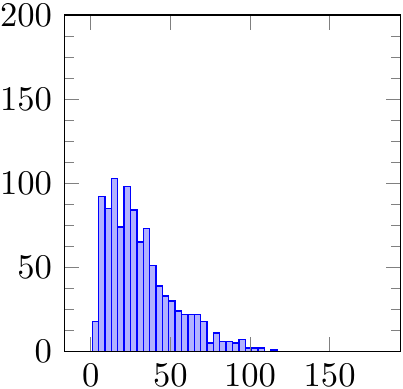}} & 
{\includegraphics[width=0.115\textwidth, height=1.6cm]{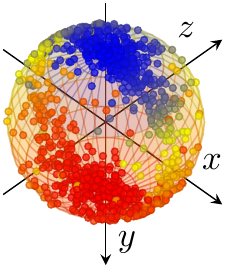}} & 
{\includegraphics[width=0.115\textwidth, height=1.44cm]{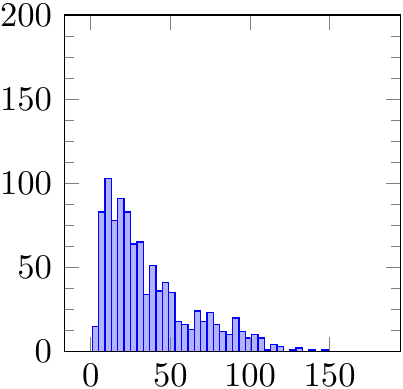}} & 
{\includegraphics[width=0.115\textwidth, height=1.6cm]{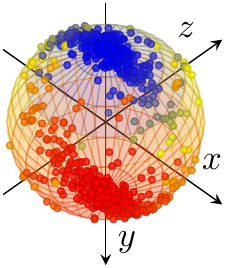}} & 
{\includegraphics[width=0.115\textwidth, height=1.44cm]{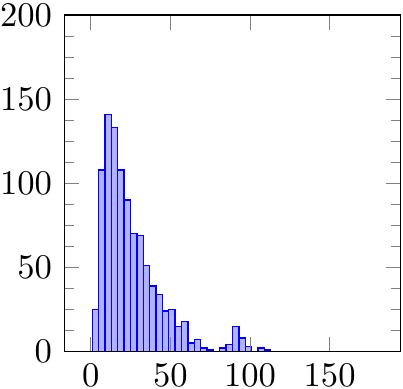}} & 
{\includegraphics[width=0.115\textwidth, height=1.6cm]{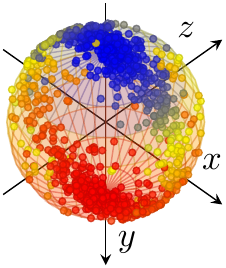}} \\

\begin{picture}(1,20)\put(-5, -12){\rotatebox{90}{~~~~~~~~~Noise~~~}}\end{picture} & 
{\includegraphics[width=0.115\textwidth, height=1.44cm]{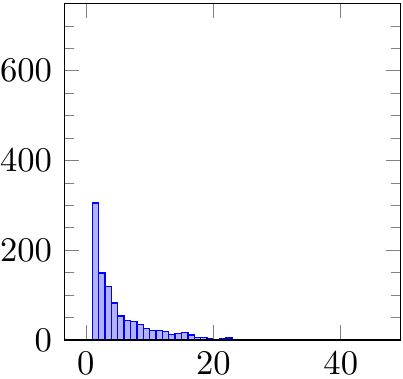}} & 
{\includegraphics[width=0.115\textwidth, height=1.6cm]{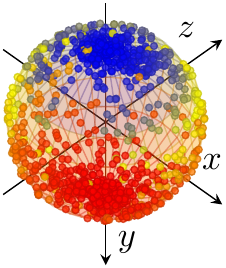}} & 
{\includegraphics[width=0.115\textwidth, height=1.44cm]{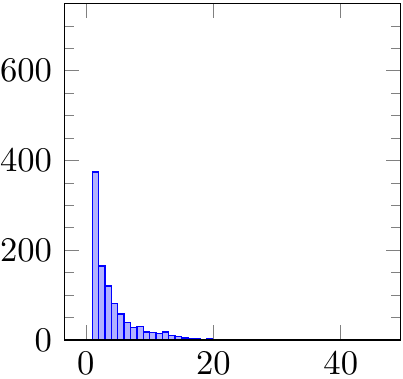}} & 
{\includegraphics[width=0.115\textwidth, height=1.6cm]{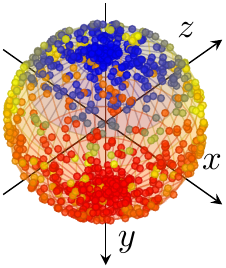}} & 
{\includegraphics[width=0.115\textwidth, height=1.44cm]{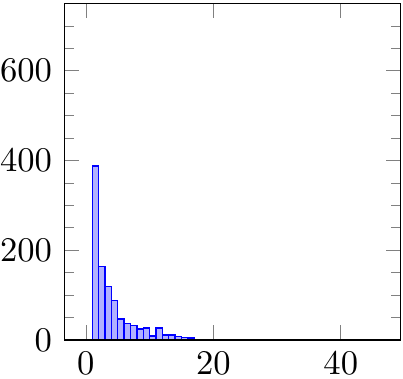}} & 
{\includegraphics[width=0.115\textwidth, height=1.6cm]{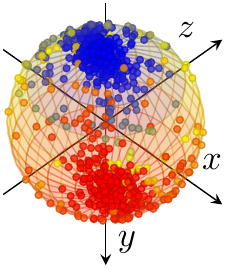}} & 
{\includegraphics[width=0.115\textwidth, height=1.44cm]{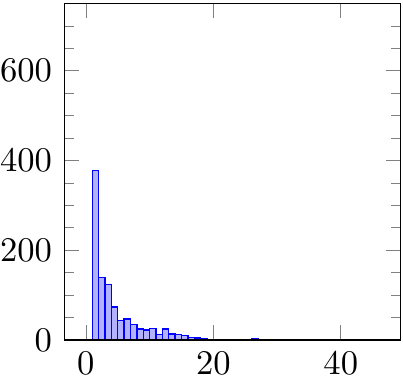}} & 
{\includegraphics[width=0.115\textwidth, height=1.6cm]{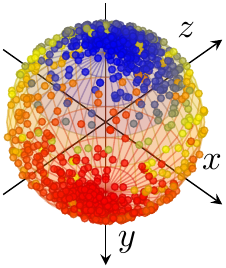}} \\
& \multicolumn{2}{c}{(i) Piccadilly}  &  \multicolumn{2}{c}{(j) Roman Forum}  &  \multicolumn{2}{c}{(k) Tower of London}  &  \multicolumn{2}{c}{(l) Trafalgar}  \\ ~\\

\begin{picture}(1,20)\put(-5, -12){\rotatebox{90}{~~~~~~~~Relative~~~}}\end{picture} & 
{\includegraphics[width=0.115\textwidth, height=1.44cm]{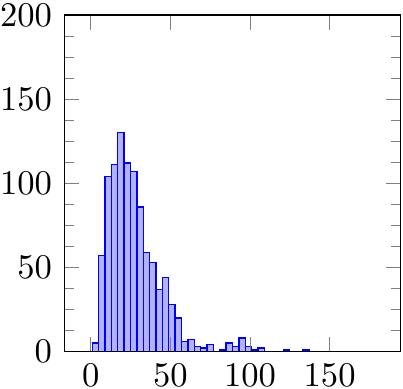}} & 
{\includegraphics[width=0.115\textwidth, height=1.6cm]{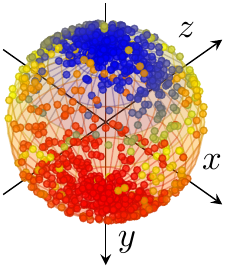}} & 
{\includegraphics[width=0.115\textwidth, height=1.44cm]{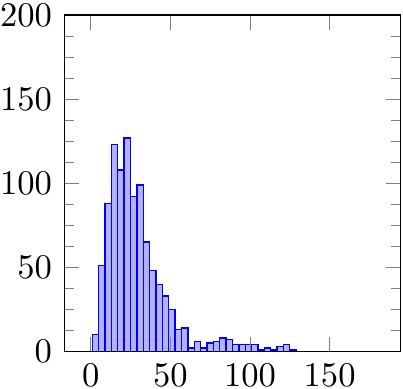}} & 
{\includegraphics[width=0.115\textwidth, height=1.6cm]{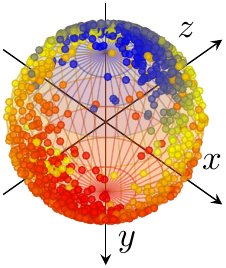}} & 
{\includegraphics[width=0.115\textwidth, height=1.44cm]{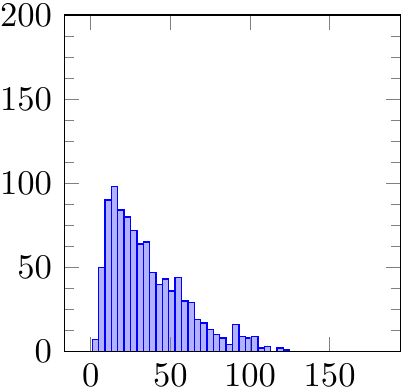}} & 
{\includegraphics[width=0.115\textwidth, height=1.6cm]{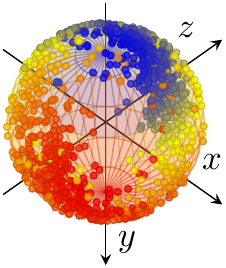}} & 
{\includegraphics[width=0.115\textwidth, height=1.44cm]{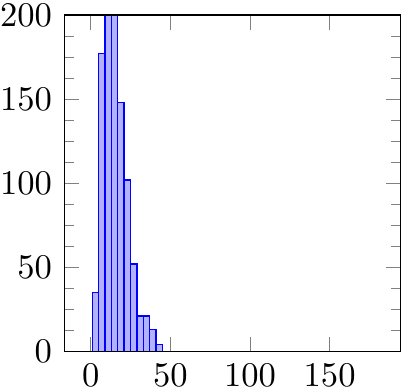}} & 
{\includegraphics[width=0.115\textwidth, height=1.6cm]{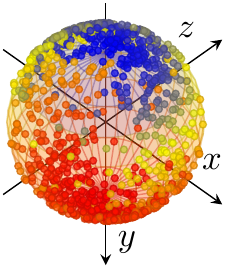}} \\

\begin{picture}(1,20)\put(-5, -12){\rotatebox{90}{~~~~~~~~~Noise~~~}}\end{picture} & 
{\includegraphics[width=0.115\textwidth, height=1.44cm]{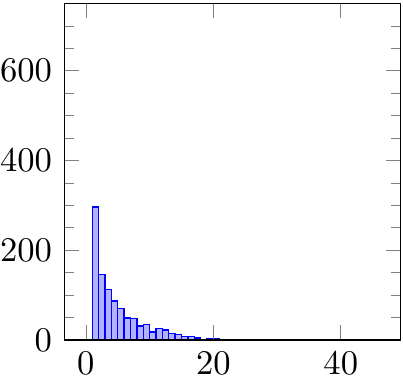}} & 
{\includegraphics[width=0.115\textwidth, height=1.6cm]{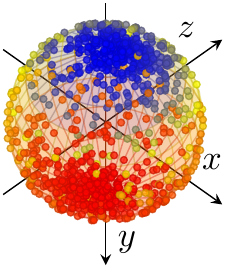}} & 
{\includegraphics[width=0.115\textwidth, height=1.44cm]{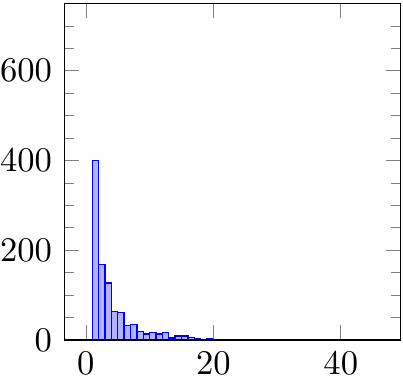}} & 
{\includegraphics[width=0.115\textwidth, height=1.6cm]{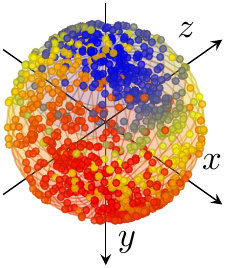}} & 
{\includegraphics[width=0.115\textwidth, height=1.44cm]{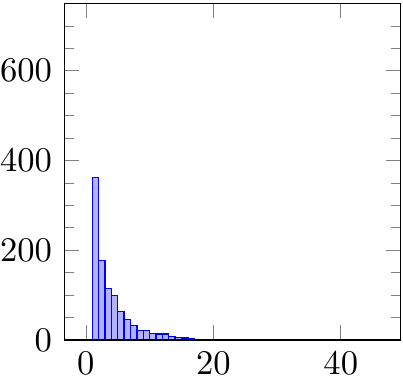}} & 
{\includegraphics[width=0.115\textwidth, height=1.6cm]{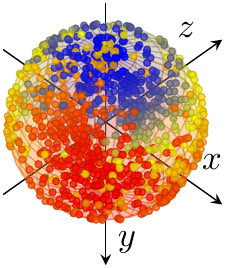}} & 
{\includegraphics[width=0.115\textwidth, height=1.44cm]{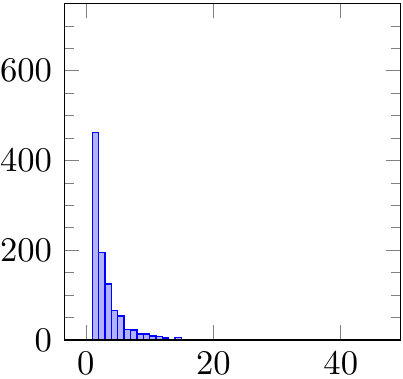}} & 
{\includegraphics[width=0.115\textwidth, height=1.6cm]{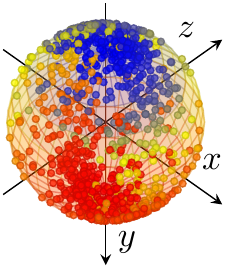}} \\
& \multicolumn{2}{c}{(m) Union Square}  &  \multicolumn{2}{c}{(n) Vienna Cathedral}  &  \multicolumn{2}{c}{(o) Yorkminster}  &  \multicolumn{2}{c}{(p) Acropolis}  \\ ~\\

\end{tabular}
 \caption[The angle and axes of observed relative orientations]{More examples of Figure~\ref{fig:statistics}(a)-(c) are displayed. The angle and axes of sampled observed relative orientations (first row) and the same of noise (second row) in real datasets (for clarity only ${1000}$ random samples) are displayed. The noise orientation is calculated from the ground-truth absolute orientations and the observed relative orientations. We plotted histograms of the magnitudes of the angles in degrees and the axes of the orientations. Notice that the axes of the sampled relative and noise orientations are distributed mostly along a vertical ring rather than uniformly on a unit ball. } 
 \label{fig:statistics}
\end{figure}
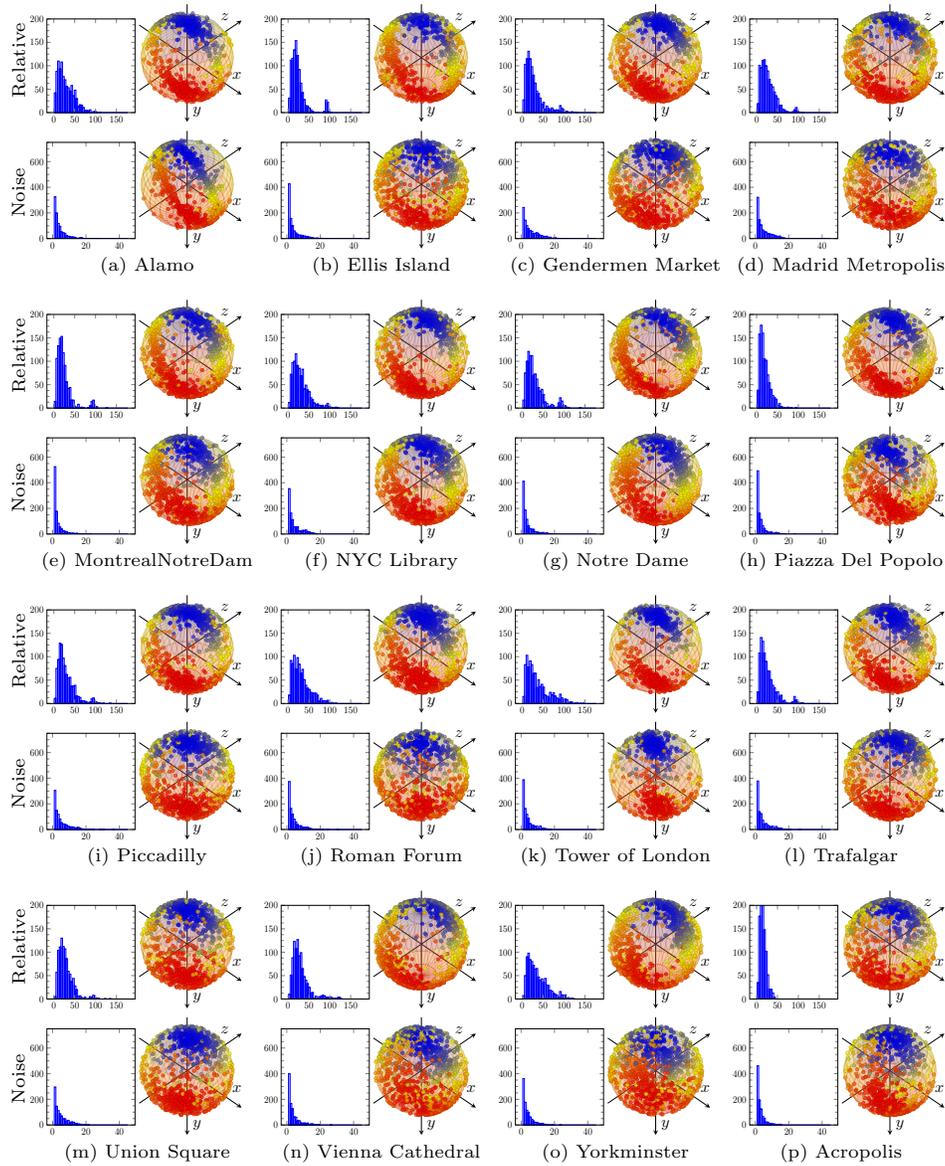


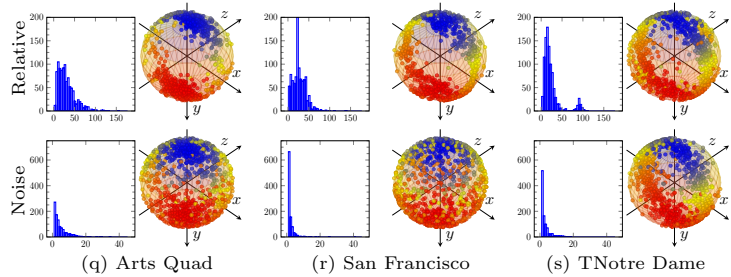
\begin{figure}
\centering \scriptsize    
\begin{tabular}{l@{\hspace{0.1em}}c@{\hspace{0.1em}}c@{\hspace{1em}}c@{\hspace{1em}}c@{\hspace{1em}}c@{\hspace{0.1em}}c} 

\begin{picture}(1,20)\put(-5, -12){\rotatebox{90}{~~~~~~~~Relative~~~}}\end{picture} & 
{\includegraphics[width=0.115\textwidth, height=1.44cm]{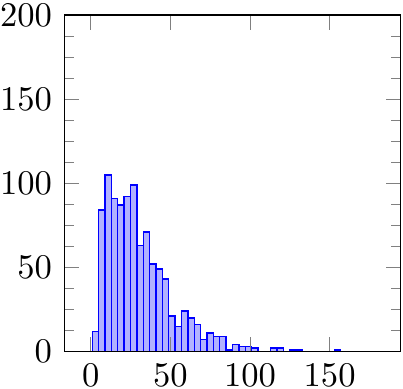}} & 
{\includegraphics[width=0.115\textwidth, height=1.6cm]{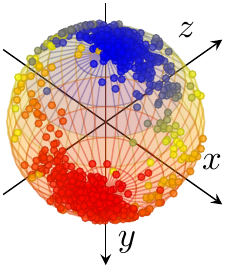}} & 
{\includegraphics[width=0.115\textwidth, height=1.44cm]{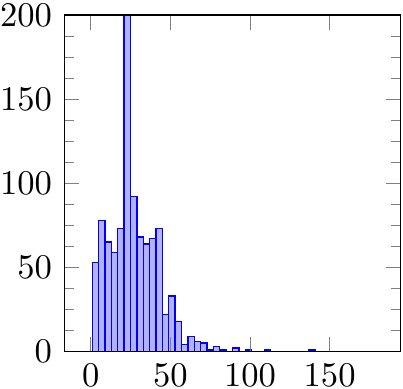}} & 
{\includegraphics[width=0.115\textwidth, height=1.6cm]{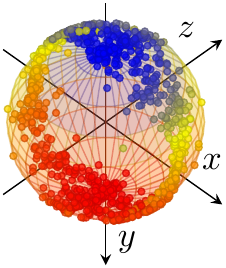}} & 
{\includegraphics[width=0.115\textwidth, height=1.44cm]{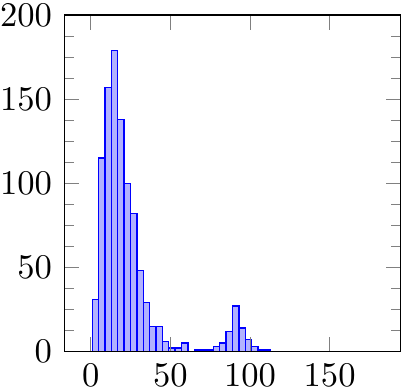}} & 
{\includegraphics[width=0.115\textwidth, height=1.6cm]{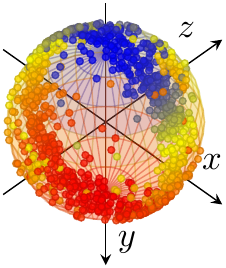}} \\ 

\begin{picture}(1,20)\put(-5, -12){\rotatebox{90}{~~~~~~~~~Noise~~~}}\end{picture} & 
{\includegraphics[width=0.115\textwidth, height=1.44cm]{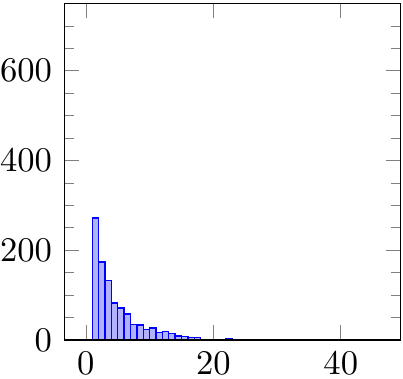}} & 
{\includegraphics[width=0.115\textwidth, height=1.6cm]{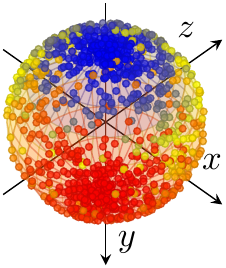}} & 
{\includegraphics[width=0.115\textwidth, height=1.44cm]{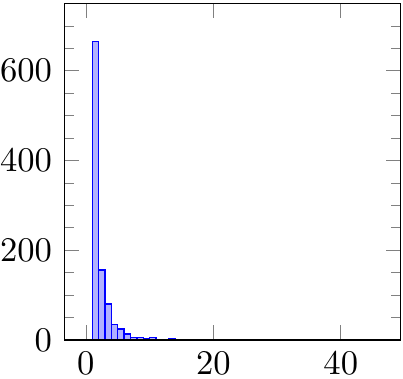}} & 
{\includegraphics[width=0.115\textwidth, height=1.6cm]{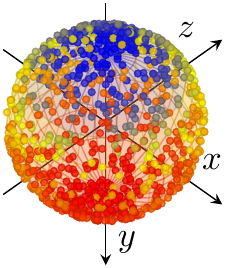}} &  
{\includegraphics[width=0.115\textwidth, height=1.44cm]{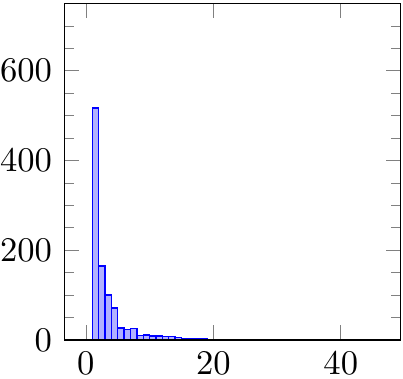}} & 
{\includegraphics[width=0.115\textwidth, height=1.6cm]{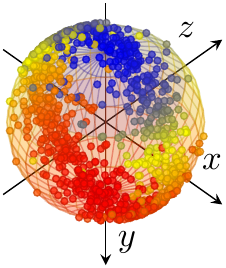}} \\  
& \multicolumn{2}{c}{(q) Arts Quad}  &  \multicolumn{2}{c}{(r) San Francisco} &  \multicolumn{2}{c}{(s) TNotre Dame}  \\ 

\end{tabular}
 \caption[The angle and axes of observed relative orientations (more examples)]{Same as Figure~\ref{fig:statistics} for the other view-graphs} 
\end{figure}
%
%

\begin{table}\setlength{\tabcolsep}{7pt}
\caption{Results of Rotation averaging on ScanNet dataset~\cite{dai2017scannet}. The average angular error on all the view-graphs in our dataset is displayed. The proposed method NeuRoRA is remarkably faster than the baselines while producing better results. Note that for these sequences, NeuRoRA takes only $\bm{0.028s}$ on average on a GPU.} 
\centering  \small   
\begin{tabular}{@{\hspace{0.1em}}p{2.3cm}@{\hspace{1em}}c@{\hspace{1em}}c@{\hspace{1em}}c|p{1.8cm}@{\hspace{1em}}c@{\hspace{1em}}c@{\hspace{1em}}c@{\hspace{0.5em}}}
\toprule
{\bf Methods} & mn & md & {cpu} &  & mn & md & {cpu} \\ 
\midrule 
{Chatterjee}~\cite{chatterjee2017robust}  &   $12.08\degree$ & $~7.80\degree$ & $~2.01$s & 
{Arrigoni}~\cite{arrigoni2018robust} &  $32.44\degree$ & $18.26\degree$ & $30.20s$   \vspace{0.25em} \\ 
{Weiszfeld}~\cite{hartley2011l1} &  $22.93\degree$ & $12.30\degree$ & $82.92s$    & 
{Wang}~\cite{wang2013exact}  &   $18.68\degree$ & $~8.53\degree$ & $~6.52$s  \vspace{0.25em} \\ 
CleanNet-SPT  &   $11.37\degree$ & $~7.41\degree$ &  $\bm{0.45s}$  &  
{NeuRoRA}  & $\bm{11.02\degree}$ & ~$\bm{6.92\degree}$ & $\bm{0.92s}$  \vspace{0.25em} \\ \midrule 
\multicolumn{8}{c}{mn: mean of the angular error; md: median of the angular error;  } \\
\multicolumn{8}{c}{cpu: the average runtime of the method;} \\ 
 \bottomrule
\end{tabular} 
\label{tab:results} 
\end{table} 

{\small
\bibliographystyle{plain}
\bibliography{egbib2}
}

\end{document}